\documentclass{article}

\usepackage{microtype}
\usepackage{graphicx}
\usepackage{booktabs} 

\usepackage[accepted]{icml2025}

\usepackage{amsmath}
\usepackage{algorithmic}
\usepackage{amssymb}
\usepackage{mathtools}
\usepackage{amsthm}

\usepackage[capitalize,noabbrev]{cleveref}

\theoremstyle{plain}

\theoremstyle{definition}

\theoremstyle{remark}

\usepackage[textsize=tiny]{todonotes}

\icmltitlerunning{Submission and Formatting Instructions for ICML 2025}

\usepackage[acronym]{glossaries}

\newacronym{dl}{DL}{Deep Learning}
\newacronym{amc}{AMC}{Automatic Modulation Classification}
\newacronym{dnn}{DNN}{deep neural network}
\newacronym{nn}{NN}{ Neural Network}
\newacronym{bn}{BN}{ Batch Normalization}
\newacronym{cnn}{CNN}{Convolutional Neural Network}
\newacronym{gap}{GAP}{Global Average Pooling}
\newacronym{fc}{FC}{Fully Connected}
\newacronym{snr}{SNR}{signal-to-noise ratio}
\newacronym{qos}{QOS}{Quality of Service}
\newacronym{5g}{5G}{Fifth Generation}
\newacronym{tl}{TL}{Transfer Learning}
\newacronym{fls}{FSL}{Few Shot Learning}
\newacronym{dg}{DG}{domain generalization}
\newacronym{tta}{TTA}{Test Time Adaptation}
\newacronym{iid}{IID}{independent and identically distributed}
\newacronym{uda}{UDA}{Unsupervised Domain Adaptation}
\newacronym{mse}{MSE}{mean squared error}

\newacronym{uav}{UAV}{unmanned autonomous vehicle}
\newacronym{xr}{XR}{extended reality}
\newacronym{vr}{VR}{virtual reality}
\newacronym{ar}{AR}{augmented reality}
\newacronym{cv}{CV}{computer vision}

\newacronym{ft}{FT}{Fine Tuning}
\newacronym{sfda}{SFDA}{Source-Free Domain Adaptation}
\newacronym{mac}{MAC}{multiply and accumulate}
\newacronym{nlp}{NLP}{natural language processing}

\newacronym{dcs}{DCS}{Discriminative Capability Scoring}

\newcommand{\beq}{\begin{equation}}
\newcommand{\eeq}{\end{equation}}
\newcommand{\bit}{\begin{itemize}}
\newcommand{\eit}{\end{itemize}}

\usepackage[framemethod=tikz]{mdframed}  

\usepackage{bm}

\usepackage{cite}
\usepackage{amsmath,amssymb,amsfonts}
\usepackage{algorithmic}
\usepackage{algorithm}

\usepackage{textcomp}
\usepackage{xcolor}
\usepackage{epstopdf}
\usepackage{graphicx}
\usepackage[utf8]{inputenc} 
\usepackage[T1]{fontenc}    
\usepackage{url}            
\usepackage{booktabs}       
\usepackage{nicefrac}       
\usepackage{microtype}      
\usepackage{xcolor}         

\usepackage{algorithm}
\usepackage{wrapfig}
\usepackage{float}
\usepackage{caption}
\usepackage{subcaption}

\usepackage{times}
\usepackage{epsfig}
\usepackage{amsmath}
\usepackage{amssymb}
\usepackage{booktabs}
\usepackage{lipsum} 
\usepackage{array}
\usepackage{multirow}
\usepackage{tabularray}
\usepackage{xspace}
\usepackage{amsthm}

\usepackage{pifont}
\usepackage{midfloat}
\newcommand{\cmark}{\ding{51}}%
\newcommand{\xmark}{\ding{55}}%

\newcommand{\dnn}{\gls{dnn}\xspace}
\newcommand{\dnns}{\glspl{dnn}\xspace}

\usepackage{natbib}

\newcommand{\FW}{SINF\xspace}

\usepackage[many]{tcolorbox}
\usepackage{lipsum}

\definecolor{titlebg}{RGB}{100,22,72}
\definecolor{introbg}{RGB}{0,128,128}

\newtcolorbox{usecase}[1][]{
  breakable,
  enhanced,
  arc=0pt,
  outer arc=0pt,
  colframe=titlebg,
  colback=titlebg!05,
  overlay unbroken and first={
    \node[
      draw=titlebg,
      fill=titlebg,
      rotate=0,
      anchor=north west,
      text=white,
      font=\bfseries
    ]
    at (frame.north west)  
    {#1};
  }
}

\newtcolorbox{mission}[1][]{
  breakable,
  enhanced,
  arc=0pt,
  outer arc=0pt,
  colframe=introbg,
  colback=introbg!05,
  overlay unbroken and first={
    \node[
      draw=introbg,
      fill=introbg,
      rotate=0,
      anchor=north west,
      text=white,
      font=\bfseries
    ]
    at (frame.north west)  
    {#1};
  }
}

\begin{document}

\twocolumn[
\icmltitle{\texttt{\FW}: Semantic Neural Network Inference with Semantic Subgraphs}

\begin{icmlauthorlist}
\icmlauthor{A.Q.M. Sazzad Sayyed}{yyy}
\icmlauthor{Francesco Restuccia}{yyy}
\end{icmlauthorlist}

\icmlaffiliation{yyy}{Northeastern University, United States}

\icmlcorrespondingauthor{A.Q.M. Sazzad Sayyed}{sayyed.a@northeastern.edu}

\icmlkeywords{Machine Learning, ICML}

\vskip 0.3in
]



\printAffiliationsAndNotice{} 

\begin{abstract}
This paper proposes \textit{Semantic Inference} (\FW)  that creates semantic subgraphs in a \gls{dnn} based on a new  Discriminative Capability Score (DCS) to drastically reduce the \gls{dnn} computational load with limited performance loss.~We evaluate the performance of \FW on VGG16, VGG19, and ResNet50 \dnns trained on CIFAR100 and a subset of the ImageNet dataset. Moreover, we compare its performance against 6 state-of-the-art pruning approaches. Our results show that (i) on average, \FW reduces the inference time of VGG16, VGG19, and ResNet50 respectively by up to 29\%, 35\%, and 15\% with only 3.75\%, 0.17\%, and 6.75\% accuracy loss for CIFAR100 while for ImageNet benchmark, the reduction in inference time is  18\%, 22\%, and 9\% for accuracy drop of 3\%, 2.5\%, and 6\%; (ii) DCS achieves respectively up to 3.65\%, 4.25\%, and 2.36\% better accuracy with VGG16, VGG19, and ResNet50 with respect to existing discriminative scores for CIFAR100 and the same for ImageNet is 8.9\%, 5.8\%, and 5.2\% respectively. Through experimental evaluation on Raspberry Pi and NVIDIA Jetson Nano, we show \FW is about 51\% and 38\% more energy efficient and takes about 25\% and 17\% less inference time than the base model for CIFAR100 and ImageNet. \vspace{-0.5cm}
\end{abstract}

\section{Introduction}
\label{sec:intro}
State-of-the-art \glspl{dnn} employ a larger number of parameters than what mobile devices can tolerate today. For example, YoLov10, the state of the art \gls{dnn} for object detection, uses a \gls{dnn} backbone with 29.5M parameters \citep{THU-MIGyolov10}. Approaches such as pruning \citep{han2015learning,chen2023a}, quantization \citep{han2015deep, qin2022bibert}, and coding \citep{gajjala2020, han2015deep} incur in excessive performance loss and most often require fine-tuning.

In this paper, we propose an approach based on \textit{cluster-level semantic DNN subgraphs} to reduce the computing load without compromising the \gls{dnn} accuracy and without retraining. As detailed in  Section \ref{sec:semantic}, semantically similar inputs share a significant number of filter activations compared to semantically dissimilar inputs. For example, as shown in Section \ref{sec:semantic}, images of seals share significantly more filter activations with images of dolphins than with images of tables. As such, if a semantic subgraph were to be available, we could only execute that subgraph related to the current input and ``turn off'' the rest of the DNN.

\begin{figure}[!ht]
    \centering
    \includegraphics[width=\linewidth]{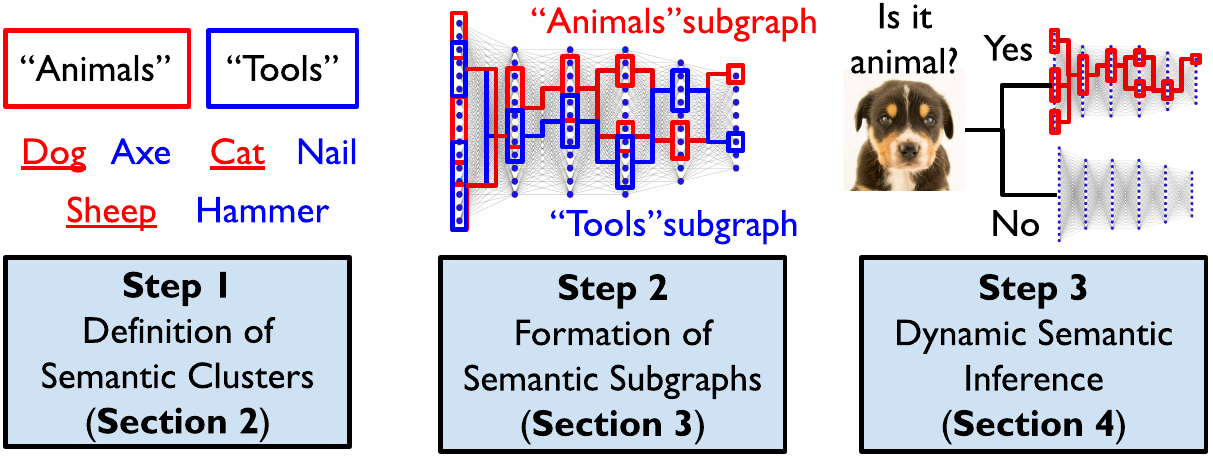}
    \caption{Overview of proposed \FW framework.}
    \label{fig:framework} \vspace{-0.2cm}
\end{figure}

We build on top of this interpretability insight and propose \textit{Semantic Inference} (\FW), whose main components are illustrated in Figure \ref{fig:framework}. In a nutshell, \FW transforms a \textit{pre-trained and static} \gls{dnn} into a \textit{dynamic} \gls{dnn} by (i) creating subgraphs corresponding to each semantic cluster, which are defined based on the current DNN task (steps 1 and 2); (ii) selecting the semantic-relevant subgraphs at inference time based on a preliminary cluster-based classification of the input image (step 3). Conversely from pruning approaches, \FW does \emph{not} require changing any portion of the DNN to decrease computing load.  \smallskip

\noindent \textbf{This paper makes the following novel contributions:} \smallskip

\noindent $\bullet$ We propose a new inference framework called \FW, which logically partitions the \gls{dnn} into semantic subgraphs based on a new \gls{dcs} that finds the filters most associated with a given semantic cluster (\textbf{Section 3}). Finally, \FW pre-classifies an image based on its relevant cluster so that only its semantic-specific subgraph is activated (\textbf{Section 4});\smallskip

\noindent $\bullet$ We benchmark the performance of \FW on VGG16, VGG19, and ResNet50 \glspl{dnn} trained on the CIFAR100 and ImageNet dataset. We compare \gls{dcs} against several state-of-the-art discriminative algorithms \citep{mittal2019,molchanov2019taylor,apoz2016,sui2021chip,lin2020hrank} (\textbf{Section 5}). Although the target of \FW is not to perform pruning but to extract subgraphs, we use DCS as a pruning approach and compare it against \citep{murti2023tvsprune} which does not require fine-tuning. We have implemented \FW on real-world platforms Raspberry Pi and NVIDIA Jetson Nano and evaluated its latency and energy consumption; \smallskip

\noindent \textbf{Key Results:}~On  average, executing the semantic subgraphs reduces the inference time of VGG16, VGG19, and ResNet50 respectively by up to 29\%, 35\%, and 15\% with only 3.75\%, 0.17\%, and 6.75\% accuracy loss for CIFAR100 while for ImageNet benchmark, the reduction in inference time is  18\%, 22\%, and 9\% for accuracy drop of 3\%, 2.5\%, and 6\%. Moreover, DCS achieves respectively up to 3.65\%, 4.25\%, and 2.36\% better accuracy with VGG16, VGG19, and ResNet50 with respect to existing discriminative scores for CIFAR100 and the same for ImageNet is 8.9\%, 5.8\%, and 5.2\% respectively. Finally, \FW is about 51\% and 38\% more energy efficient and takes about 25\% and 17\% less inference time than the base DNNs for CIFAR100 and ImageNet, respectively.

\section{Related Work}
\label{sec:related_work}

Table \ref{tab:difference_of_work} compares \FW with prior approaches, which are detailed in the rest of the section. Unlike early exit methods, which require multiple classifier heads, \FW requires a single auxiliary classifier. As for pruning and quantization, these are considered orthogonal to \FW and can be used in addition to \FW to further improve performance.

\begin{table}[!h]
\centering
\begin{tabular}{|c|c|c|c|}
\hline
             & \textbf{Dynamic} & \textbf{Fine-tuning} & \textbf{Semantic}  \\ \hline
\textbf{Pruning}      & \xmark       & \cmark                   & \xmark \\ 
\textbf{Quantization} & \xmark       & \cmark                   & \xmark \\ 
\textbf{Early-exit}   & \cmark       & N/A                 & \xmark \\
\textbf{\FW}         & \cmark       & \xmark                   & \cmark\\     
\hline
\end{tabular}
\caption{\FW vs other related approaches. \vspace{-0.3cm}}
\label{tab:difference_of_work}
\end{table}

\noindent \textbf{Pruning:}~Han et al.~\citep{han2015learning} proposed weight-norm-based unstructured pruning. Li et al.~\citep{li2017pruning} used the L\textsubscript{1} norm of the kernel weights to prune entire filters. However, weight-norm-based strategies do not directly take into account the importance of the filters or parameters to preserve the \gls{dnn} accuracy. Another approach employs first-order gradient which estimates the importance of the filters based on the gradient of the loss function \citep{molchanov2019taylor}. Another class of techniques find and prune duplicate or redundant filters. To find such filters, Sui et al.~\citep{sui2021chip} use the change in the nuclear norm of the matrix formed from the activation maps when individual filters are removed from a layer. Lin et al.~\citep{lin2020hrank} use the expected rank of the feature maps, while Chen et al.~ \citep{chen2023a} explain the soft-threshold pruning as an implicit case of Iterative Shrinkage-Thresholding. \textit{Although these methods determine the redundant filters, they fail to focus on the filters that are necessary to distinguish among the classes. Moreover, all of these methods require fine-tuning after pruning}. When fine-tuning is not possible, these methods do not provide satisfactory performance. Recently, Murti et al.~\citep{murti2023tvsprune} proposed a retrain-free \emph{IterTVSPrune} approach based on Total Variational Distance (TVD) \citep{verdu2014total}. Here, we extract subgraphs most representative of classes belonging to a given semantic cluster.\smallskip

\noindent \textbf{Quantization and Coding}:~The seminal work by \citep{han2015deep} compressed the \gls{dnn} through quantization and Huffman coding to reduce the memory footprint. Among more recent work, post-training quantization \citep{cai2020zeroq} and quantization-aware training \citep{Zhong2022} have been proposed. Qin et al.~\citep{qin2022bibert} pushes the boundary using single-bit quantization of the popular language model Bidirectional Encoder Representations from Transformers (BERT). Lin et al.~\citep{li2020pams} designed a layer-wise symmetric quantizer with the learnable clip value only for high-level feature extraction module. Tu et al.~\citep{tu2023toward} recently designed an algorithm for network quantization catered to the needs of image super-resolution. Both quantization and coding are orthogonal to \FW and can be used to achieve further improvement. \smallskip

\noindent \textbf{Early Exiting}:~Early exiting was proposed by \citep{teerapittayanonbranchynet} to make the \gls{dnn} inference dynamic by using auxiliary (and relatively small) classifiers attached to the output of the \gls{dnn} layers \citep{matsubara2021split}. Based on their confidence, the decision to traverse the remaining layers is made \citep{matsubara2021split}. The training of the auxiliary classifiers can be done jointly with the backbone network \citep{pomponi2022probabilistic} or separately \citep{garg2021will}. The classifiers can be trained using either cross-entropy loss \citep{Wang2019DyNExit}, or knowledge distillation \citep{Phuong2019DistEE}. Recently, \citep{han2023blockloss} proposed using block-dependent loss from a subset of the exits close to a block to train the classifiers. Dong et al.~\citep{Dong2022EP} predicts which early exit to use using a lightweight "Exit Predictor". \citep{Hari2023unsupervisedEE} modeled the exit selection as an online learning problem and  chooses the exit in an unsupervised way. Conversely, SINF uses an auxiliary classifier to select the subgraph according to the input. \vspace{-0.3cm}

\section{Dividing a DNN into Semantic Subgraphs}\label{sec:semantic}

We define the concept of \textit{semantic cluster}. Let $\mathcal{D}$ be a labeled dataset with class labels $\mathcal{K}$.~We define $K$ semantic clusters, each composed by a subset of classes $\{\gamma_1, \dots, \gamma_{K}\}$ such that $\gamma_1 \cup \gamma_2 \cup \cdots \gamma_{K} = \mathcal{K}$. We primarily assume that these clusters are formed based on similarity of the semantics of their member classes. These semantics can be defined at the application level. The clusters can also be pre-defined at the dataset level (e.g., as in the CIFAR100 dataset). 

\begin{figure}[!h]
    \centering
    \includegraphics[width=\linewidth]{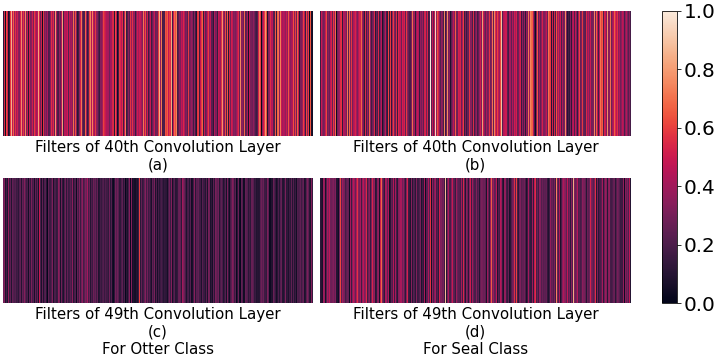}\\
    \includegraphics[width=0.8\linewidth]{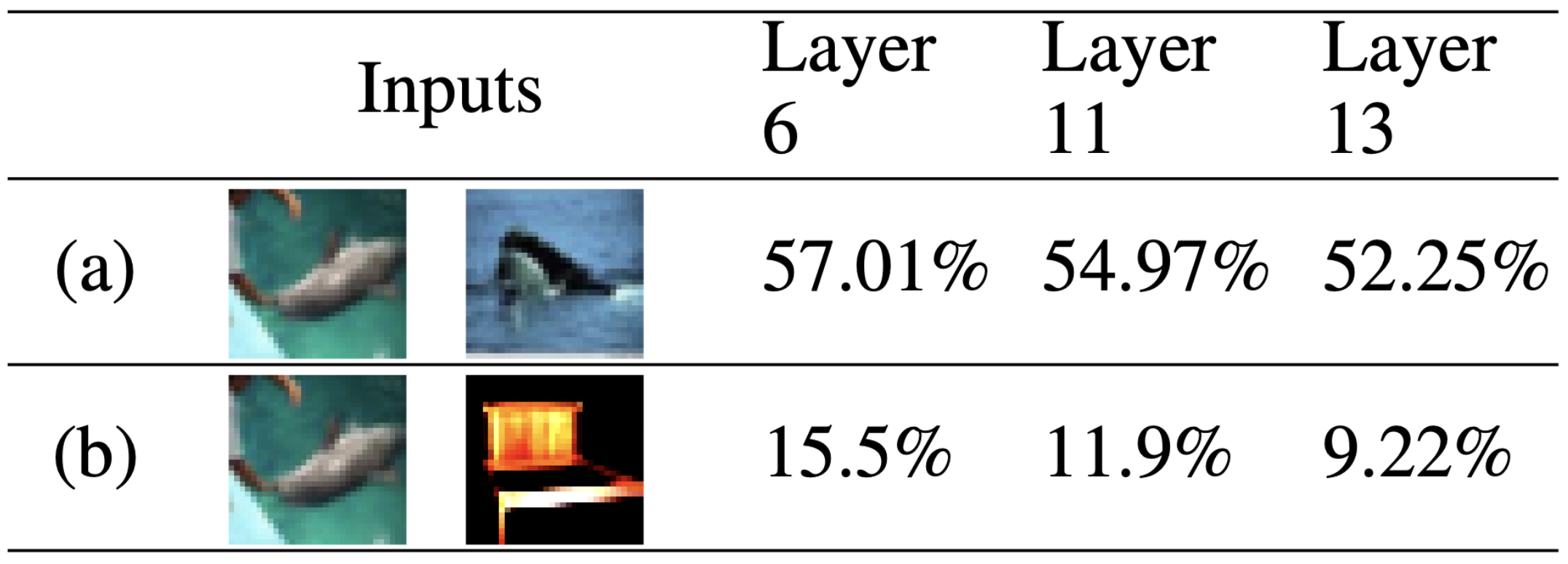}
    \caption{(top) Filter activations in ResNet50; (bottom) Percentage of filters shared between (a) semantically similar ``dolphin'' and ``whale''; (b) semantically dissimilar  ``dolphin'' and ``table''. \vspace{-0.3cm}} 
    \label{fig:activation_pattern}
\end{figure}

We performed a series of experiments to validate the intuition behind our proposed \FW approach (we refer to supplementary section \ref{sec:motivating_exp_1} for the details).~It is well known that filters of \gls{dnn} identify parts of objects, colors or concepts.~Many of these filters are shared among classes \citep{bau2017networkdissection}.~On the other hand, filter activations become sparser as the \gls{dnn} becomes deeper, with filters reacting only to specific inputs  belonging to specific classes.~This phenomenon can be observed in the top portion of \ref{fig:activation_pattern}, which  shows the average filter activation strength for the ``otter'' and ``seal'' classes in the the 40th and 49th convolutional layers of ResNet50  trained on CIFAR100.

\textbf{Key Observations.}~\textit{This experiment indicates that filters in earlier layers are less specialized than filters in deeper layers}. Moreover, it remarks that filters from semantically similar classes get similarly activated, especially in earlier layers. \textit{To put it in more quantitative terms, the L\textsubscript{1} distance of the activation maps of the mentioned classes in the 40\textsuperscript{th} layer is 0.028, while the same for the 49\textsuperscript{th} layer is 0.111}. To further investigate this critical aspect, we have performed additional experiments where we have computed the percentage of filters ``shared'' among different classes for each layer of VGG16. Specifically, we have tagged each filter with the top 20 classes for which it gets activated. For each pair of classes, their similarity is calculated as the number of filters tagged with both classes over the number of filters tagged with at least one of the classes. The results are shown in the bottom portion of Figure \ref{fig:activation_pattern}, where the first row shows the filters shared between the ``dolphin'' and ``whale'' classes -- two semantically similar classes. The second row shows the filter sharing between two semantically dissimilar classes - ``dolphin'' and ``table''. \textit{We notice that the semantically similar classes share more filters}. 

A key issue is to define the subgraphs corresponding to each semantic cluster. We formalize this step as Semantic DNN Subgraph Problem (SDSP). We consider a \gls{dnn} $\mathcal{F}$ trained on dataset $\mathcal{D}$ as a computation graph, while the filters of the \gls{dnn} work as the nodes of the graph.

\begin{mission}[Semantic DNN Subgraph Problem (SDSP)]
\vspace{0.2in}
Find $K$ subgraphs $\mathcal{F}_{\gamma_i} \cdots \mathcal{F}_{\gamma_K}$   such that

\begin{equation} \label{eq:problem_formulation}
    \mathcal{B}_{eval}(\mathcal{F}, \mathcal{D}_{\gamma_i}) \leq \mathcal{B}_{eval}(\mathcal{F}_{\gamma_i}, \mathcal{D}_{\gamma_i}) + \epsilon,
\end{equation}
where $\epsilon$ is an error margin and  $\mathcal{F}_{\gamma_i} \subset \mathcal{F}$ and $\mathcal{D}_{\gamma_i} \subset \mathcal{D}$ are respectively the proper subgraphs of $\mathcal{F}$ and subset of data corresponding to the semantic cluster $\gamma_i$. The function $\mathcal{B}_{eval}$ is the metric to measure the performance of the \gls{dnn} on the subsets of dataset corresponding to semantic clusters. A higher value of $\mathcal{B}_{eval}$ corresponds to better performance. 
\end{mission}

In other words, the subgraph $\mathcal{F}_{\gamma_i}$ contains the nodes of $\mathcal{F}$ which best classifies the members of the semantic cluster $\gamma_i$ within the error margin of $\epsilon$. \vspace{-0.2cm}

\section{Discriminative Capability Score}

We design a novel algorithm named Discriminative Capability Score (\gls{dcs}), which aims at satisfying Equation \ref{eq:problem_formulation} by extracting the filters from each layer of a \gls{dnn} which best discriminate among the members of a semantic cluster $\gamma$. We describe the \gls{dcs} for a given layer $l$ and given semantic cluster $\gamma_m$ in algorithm \ref{alg:dcs} (in Appendix). We start by considering the activation map $\mathbf{A}_l^j \in \mathbb{R}^{C_{out}^l \times k \times k}$ of a generic layer $l$ of a \gls{dnn} for input $X^j (\mbox{with target label}\ t^j) \in \mathcal{D}_{\gamma}$. Here, $C_{out}^l, \mbox{and}\ k$ is the number of channels and size of a single channel of the activation map. We apply an adaptive average pooling operation $\mathcal{P}$ to obtain $\Tilde{A}_l^j \in \mathbb{R}^{C_{out}^l \times k' \times k'}$, where $k'$ is the reduced size of a single channel of the activation map. We then flatten the activation map to obtain feature map $\mathbf{F_l^j} \in \mathbb{R}^{C_{out}^lk'^2}$ for the layer $l$ and input $X^j$. Our goal is to first learn a linear transformation $\mathbf{W}_l \in \mathbb{R}^{|\gamma| \times C_{out}^lk'^2}$ ($|\gamma| = \mbox{cardinality of set}\ \gamma$) that can distinguish the members of $\gamma$ from the feature maps $\mathbf{F}_l^j$. To this end, we minimize the objective function $\mathcal{L}_{DOF}$: \vspace{-0.3cm}

\begin{equation} \label{eq:dfs_objective}
    \mathbf{W}^*_l = \underset{\mathbf{W}} {\mathrm{argmin}}\  \frac{1}{|\mathcal{D}_{\gamma_m}|}\sum_{j=1}^{j=|\mathcal{D}_{\gamma_m}|} \mathcal{L}_{DOF} (\mathbf{W}_l \cdot \mathbf{F}^j_l, t^j),
\end{equation}

Once the transformation $\mathbf{W}_l$ is learned, the importance of the features and in turn the filters, are encoded in $\mathbf{W}_l$. 

\begin{figure}[!h]
    \centering
    \includegraphics[width=0.95\linewidth]{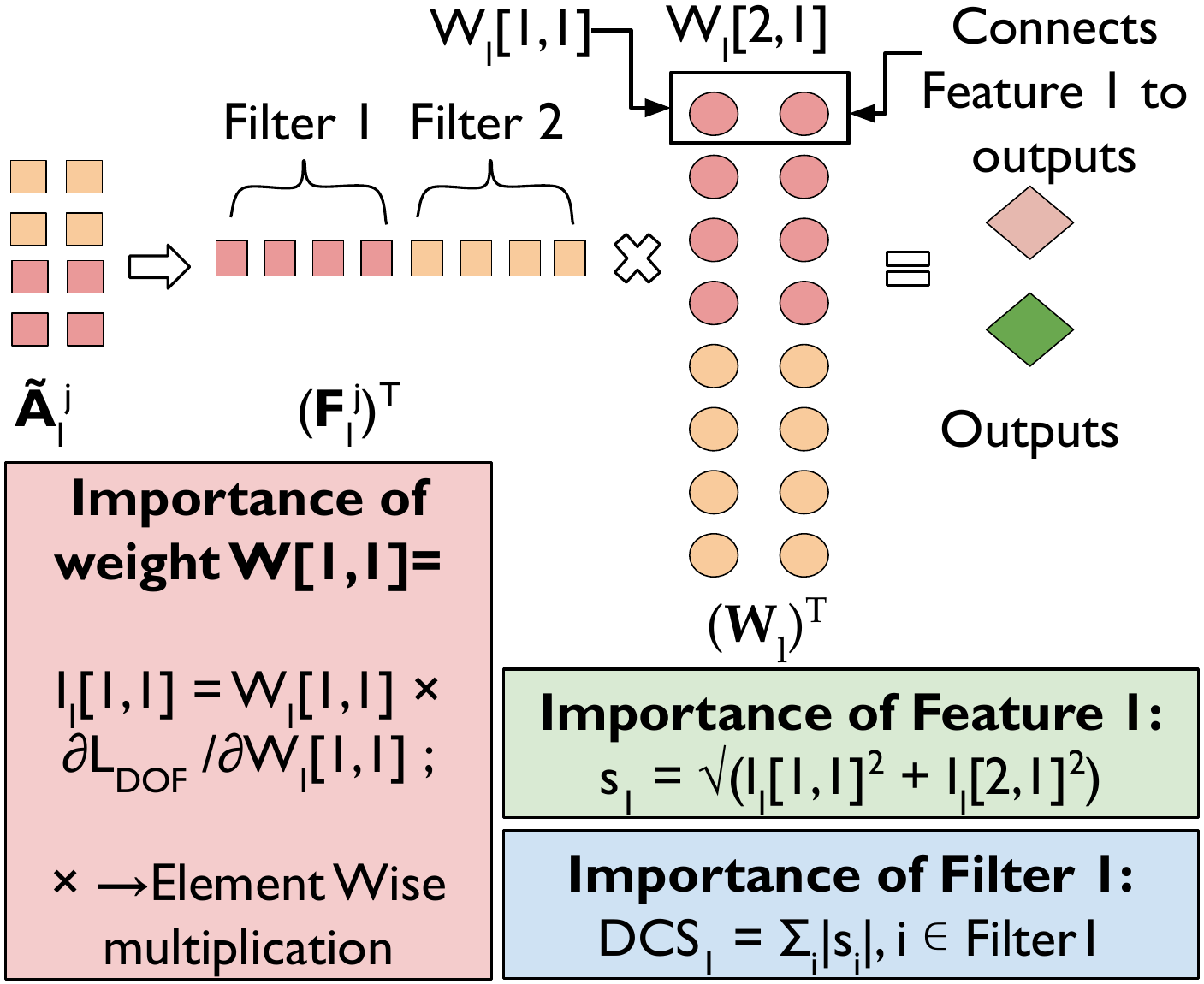}
    \caption{Example of computation of \gls{dcs} score. \vspace{-0.3cm}}
    \label{fig:dcs_explanation}
\end{figure}

We provide an example of DCA in Figure \ref{fig:dcs_explanation}. For simplicity, we show the feature vector and the weight matrix in transposed form. Notice that each column of the weight matrix $\mathbf{W}_l$ connects a single feature to the outputs. The weights of these connections can be used to directly measure the importance of the feature. The importance of the $i$-th feature in discriminating among the members of the cluster depends on the weight of its connections to the outputs and on the sensitivity of those weights, i.e., the gradient of the objective function with respect of those weights. We formalize this notion in Proposition \ref{prop:1}. 

As a result, the importance of i-th feature can be calculated as $s_i = \sqrt{\sum_{j=1}^{C_{out}^lk'^2}\mathbf{I}[ j, i ]^2}$. As $k'^2$ consecutive features come from the same filter, we calculate the DCS of the $i$-th filter of the $l$-th layer as $\mbox{DCS}_i^l = \sqrt{\sum_j |s_j|}, j \in \mbox{Filter i}$, where j denotes indices of the features that come from the $i$-th filter.     



\begin{figure}
\centering
     \begin{subfigure}{0.4\linewidth}
         \centering
         \includegraphics[width=\linewidth]{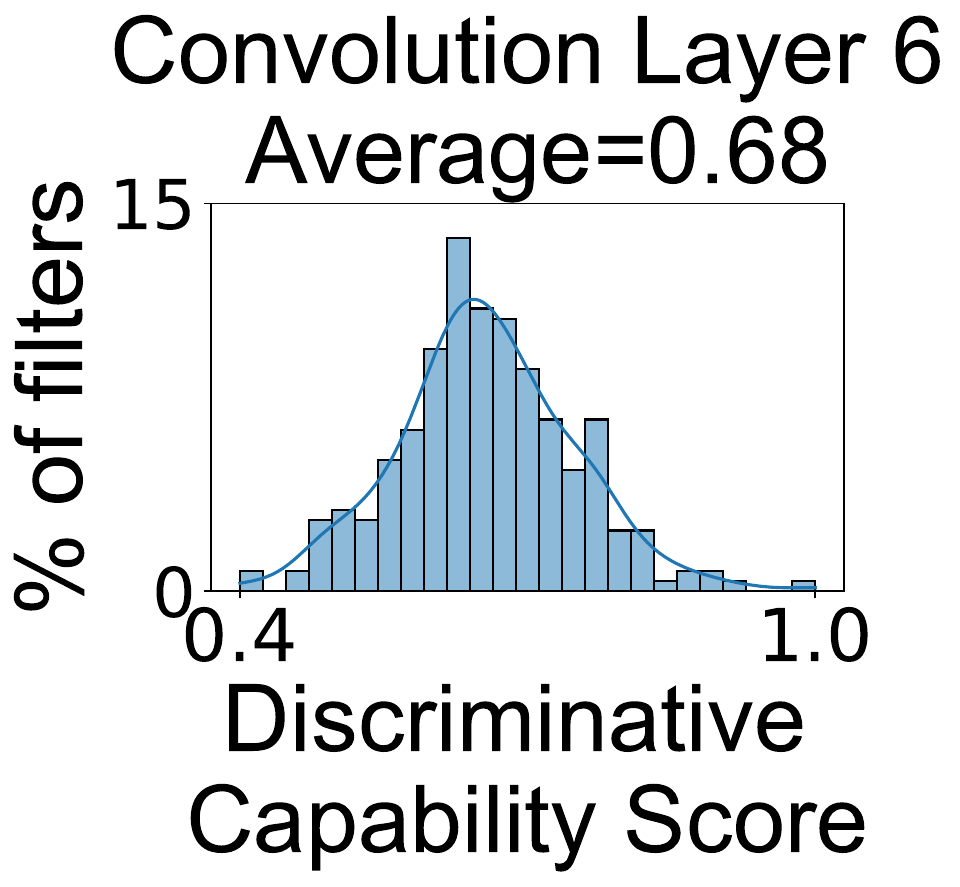}
         \caption{}
         \label{subfig:DCSdist6}
     \end{subfigure}
     \begin{subfigure}{0.4\linewidth}
         \centering
         \includegraphics[width=\linewidth]{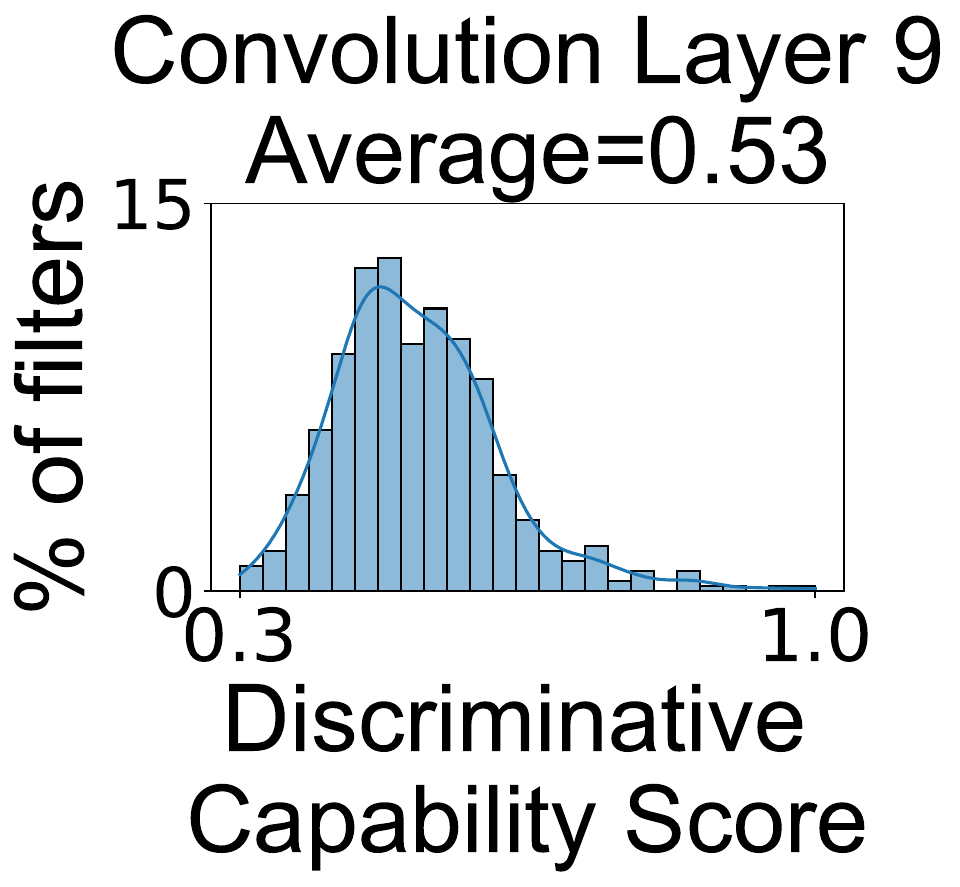}
         \caption{}
         \label{subfig:DSCdist9}
     \end{subfigure}

     \begin{subfigure}{0.4\linewidth}
         \centering
         \includegraphics[width=\linewidth]{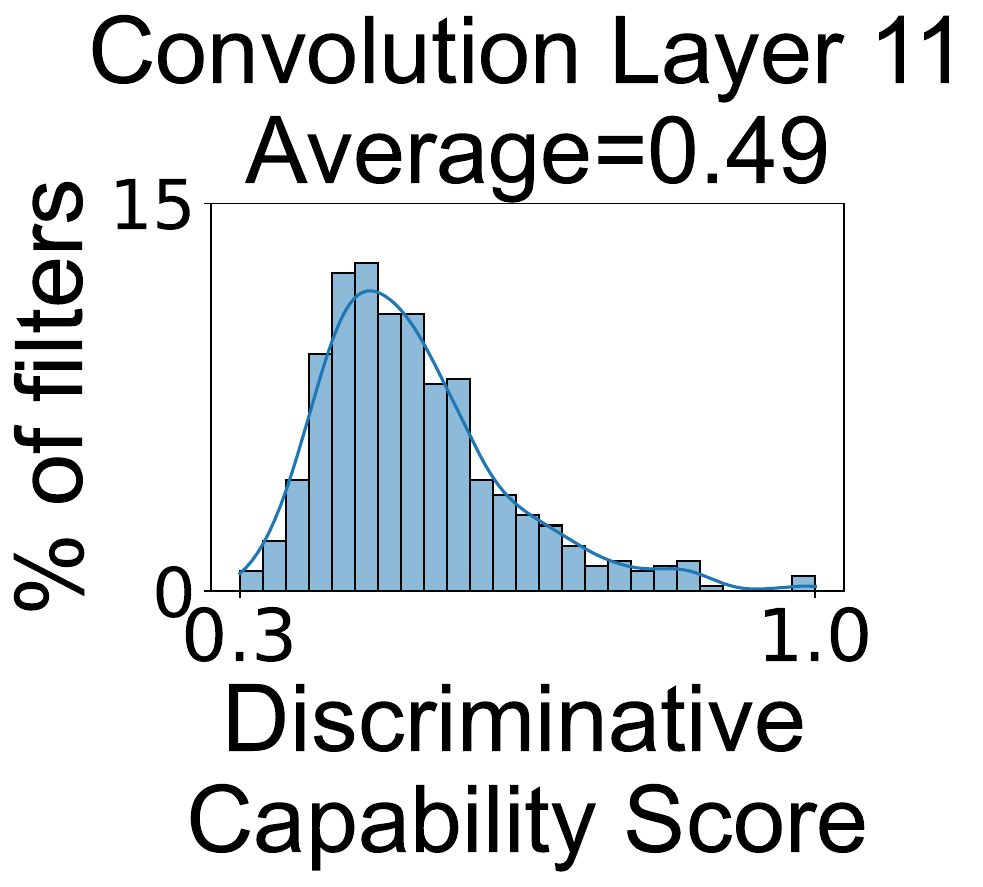}
         \caption{}
         \label{subfig:DSCdist11}
     \end{subfigure}
     \begin{subfigure}{0.35\linewidth}
         \centering
         \includegraphics[width=\linewidth]{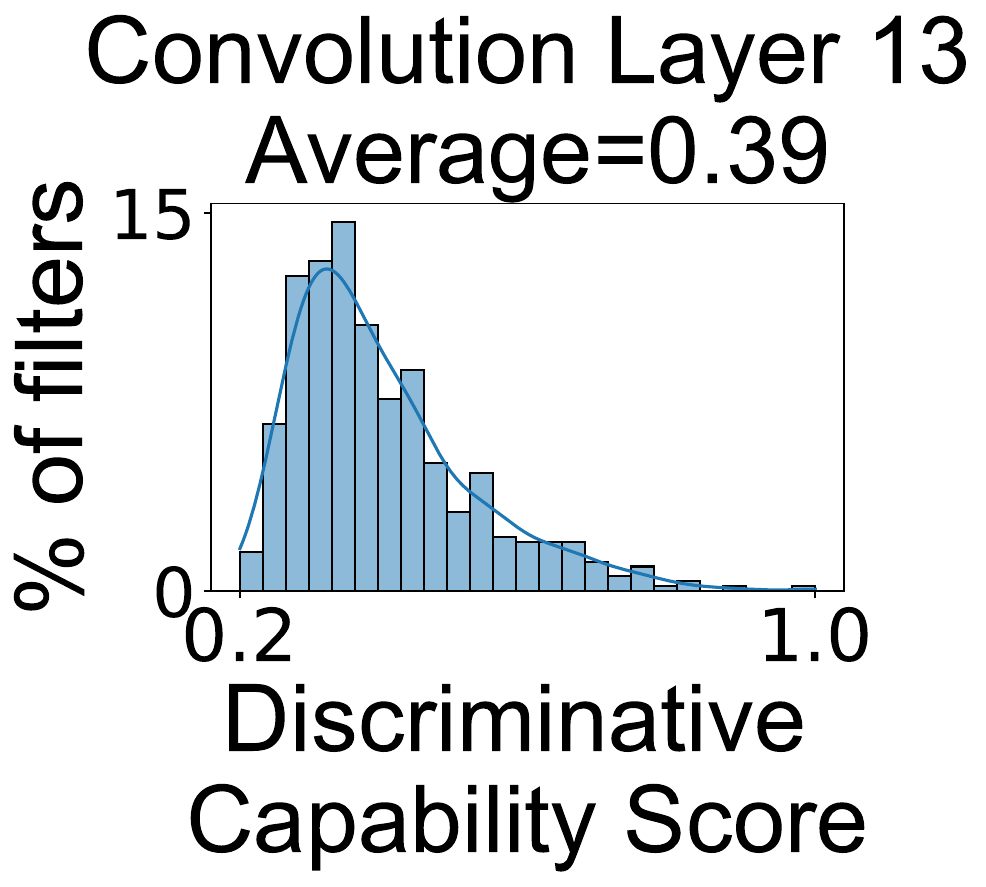}
         \caption{}
         \label{subfig:DSCdist13}
     \end{subfigure}
     \caption{The DCS distribution for cluster ``fish'' of CIFAR100 in different layers of VGG16.}
     \vspace{-0.4cm}
     \label{fig:DSCdsit}
\end{figure}

\noindent Figure \ref{fig:DSCdsit} shows the \gls{dcs} distributions obtained in layers 6, 9, 11, and 13 of VGG16 by considering the cluster ``fish'' of CIFAR100. Figure \ref{fig:DSCdsit} confirms that deeper layers are more specialized for individual classes, and thus the average \gls{dcs} for the filters in the deeper layers is smaller, i.e., 0.68 for layer 6 vs 0.39 for layer 13.  As such, \gls{dcs} captures the filter activation pattern of the \dnn.

\captionsetup[subfigure]{labelformat=empty}

\section{Dynamic Semantic Inference (SINF)}

\begin{figure}[!ht]
    \vspace{-0.3cm}
    \centering
    \includegraphics[width=0.9\linewidth]{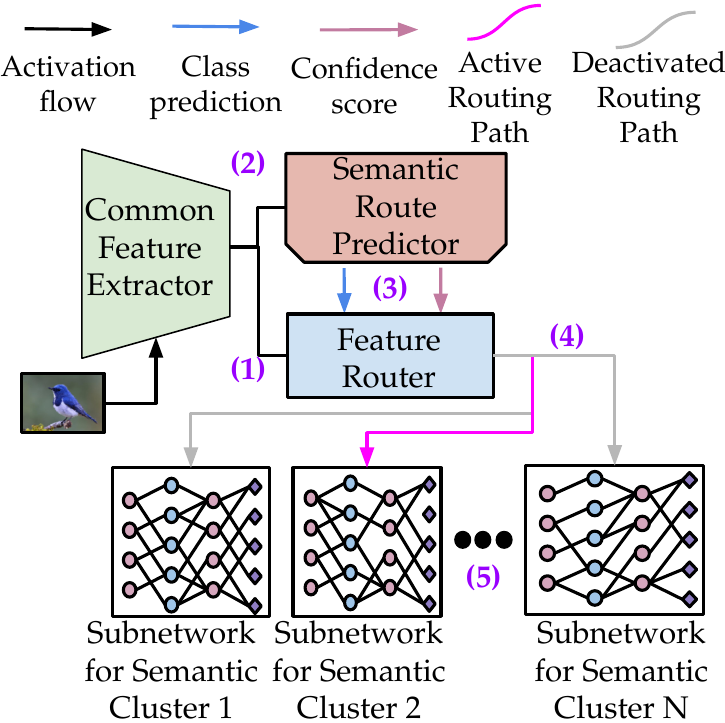}
    \caption{Overview of SINF.\vspace{-0.3cm}}
    \label{fig:overview}  
\end{figure}

\noindent A step-by-step overview of the main operations of \FW is summarized in the top portion of Figure \ref{fig:overview}.~A key challenge is assigning each incoming input to a semantic cluster. For this reason, we divide the DNN into two portions -- a \textit{Common Feature Extractor} (CFE) and the \emph{Semantic Subnetworks} (SSN). The output of the CFE is used by a \textit{Semantic Route Predictor} (SRP) that assigns the input to a semantic cluster (\textbf{step 1}). To this end, the features extracted by the CFE are passed to the SRP (\textbf{step 2}). The SRP, detailed later in this section, provides both the predicted semantic cluster and its confidence in its prediction to the \emph{Feature Router} (FR) (\textbf{step 3}). Based on the SRP output, the features extracted by the CFE will be routed to the selected semantic subgraph by using the FR (\textbf{step 4}). Finally, the extracted subgraphs predict the output (\textbf{{step 5}}).
Although Figure \ref{fig:overview} represents each subgraph separately for better graphical clarity, no additional memory beyond the annotations is needed to characterize each subgraph.  \smallskip

\noindent \textbf{Semantic Route Predictor.}~The SRP predicts the semantic clusters using an auxiliary classifier $\boldsymbol{\chi}$, which is attached after the $M-1$-th layer of $\mathcal{F}$. In our experiments, we chose as $M$ the earliest layer providing classification accuracy of 75\%. As such, the layers of $\mathcal{F}$ up to the $M-1$-th layer become the CFE. In our current design, the architecture of the auxiliary classifier consists of two convolutional layers, followed by an adaptive average pooling layer stacked on top of three fully connected layers. We use the convolutional layers to tailor the activation map from layer $l$ of $\mathcal{F}$ for the classification of the semantic clusters. To train the auxiliary classifier $\boldsymbol{\chi}$, the first $M-1$ layers of $\mathcal{F}$ are frozen and the classifier is trained in supervised fashion using $\{\mathbf{A}_{M-1}^j, \gamma_m^j\}_{j=1}^{j=|\mathcal{D}|}$ as the dataset. Here, $\mathbf{A}_{M-1}^j$, and $\gamma_m^j$ are respectively the activation of the $M-1$-th layer of the $\mathcal{F}$ and the ground truth semantic cluster for the $j$-th sample. As we are considering a pre-trained DNN, we train the auxiliary classifier separately from the DNN using the activations obtained from the $M-1$-th layer.~The output of the SRP is the probability distribution over the $K$ different semantic clusters, and the input is assigned to the semantic cluster with the highest probability. 

\begin{figure}
    \centering
    \includegraphics[width=1.2\linewidth, angle=270]{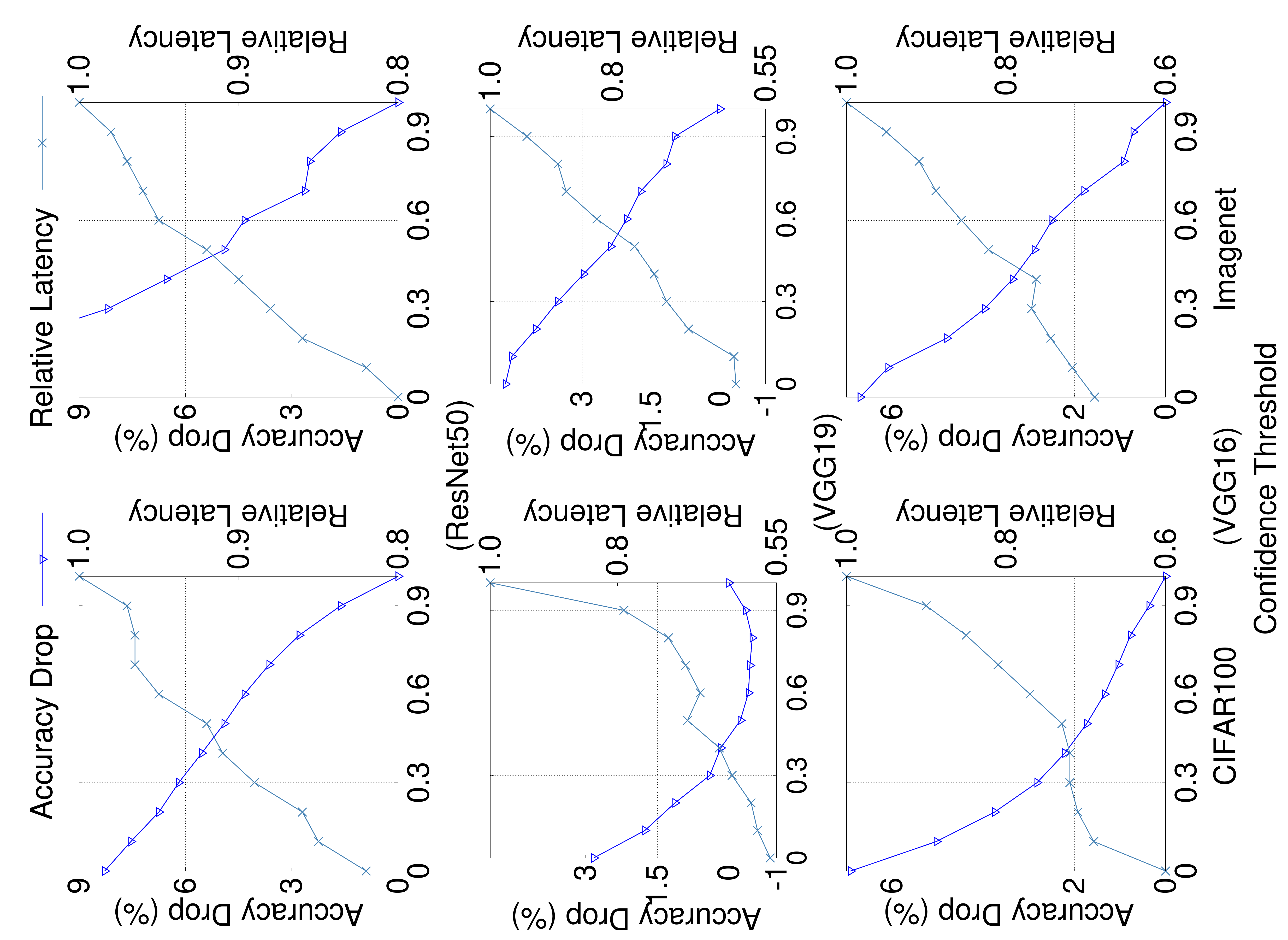}
    \caption{Accuracy and inference time vs confidence threshold for CIFAR100 (left) and ImageNet (right). \vspace{-0.5cm}}
    \label{fig:model_conf_vals_acc_inference}
\end{figure}



\noindent \textbf{Extraction of Subgraphs.} The extraction of the subgraph follows the procedure described in Algorithm \ref{alg:subgraph_extraction} in Appendix \ref{sec:subgraph_extract}. We define $L$ and $M$ as respectively the last layer of the base model $\mathcal{F}$ and the layer after the CFE. We define $r_{l}$ as the percentage of retained filters in generic layer $l$. For semantic cluster $\gamma_i$, we iterate from layer $L$ to layer $M$ to extract the subgraph. For each layer $M \le l \le L$, we calculate $r_l (r_L \leq r_l \leq r_M)$, as well as the DCS score of the filters using \gls{dcs} algorithm. For each cluster, we rank the filters based on the \gls{dcs} score, and the indices of the top $r_l$\ percent filters are saved. If the average accuracy of the subgraphs is above threshold $\tau_{acc}$, the indexes of the filters belonging to the subgraphs are stored. This procedure is performed for different values of $r_L$ and $r_{M}$. Further details are provided in Section \ref{sec:exp-set}. \smallskip

\noindent \textbf{Feature Router.}~The effectiveness of the DCS score can be improved by conditioning the outputs of the SRP to the confidence of the SRP $\boldsymbol{\chi}$. A higher confidence value represents a higher probability that the SRP is able to correctly place the input in the proper semantic cluster. The Feature Router (FR) calculates this confidence by taking the activation map from $\boldsymbol{\chi}$ along with the probability distribution from its prediction layer. To compute the confidence of the classifier on individual decisions, the FR employs the lightweight metric in \citep{BLDNN2015}. The confidence score can be calculated as $C_{\chi} = P_{h} - P_{sh}$,  using the highest ($P_h$) and the second highest probabilities ($P_{sh}$) for individual semantic clusters. If the confidence score exceeds a threshold, the activation map is routed to the subgraph corresponding to the predicted semantic cluster. Otherwise,  $\mathcal{F}$ is fully executed to obtain the inference task output. \vspace{-0.2cm}




\begin{figure}
    \centering
    \includegraphics[width=1.2\linewidth, angle=270]{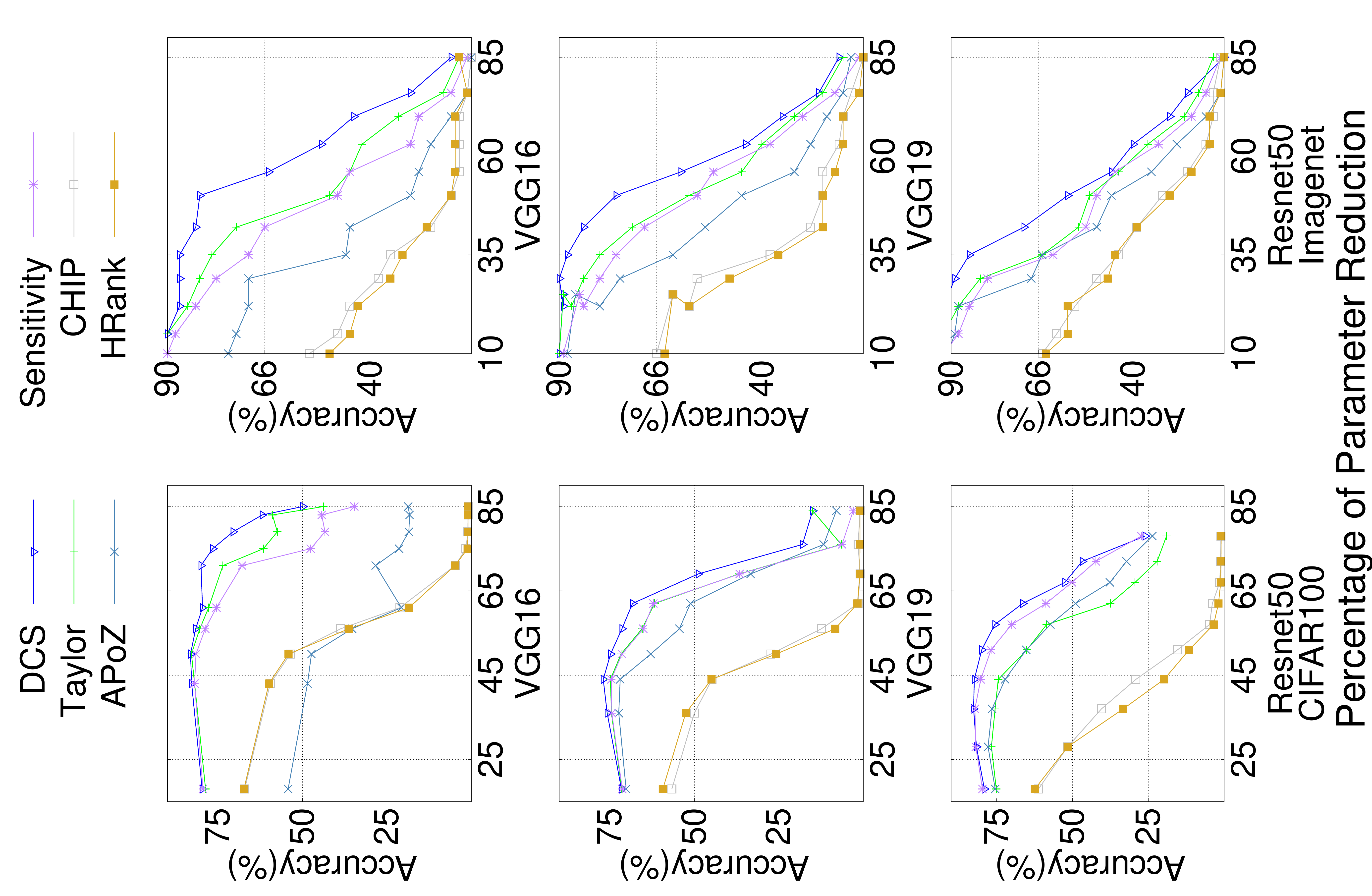}
    \caption{Comparison between DCS and other parameter reduction approaches on CIFAR100 (left) and ImageNet (right). \vspace{-0.6cm}}
    \label{fig:comp_acc}
\end{figure}

\section{Experimental Results}
We perform an extensive set of experiments to understand the utility of \FW. We defer the experimental setup to supplementary section \ref{sec:exp-set} and discuss the experimental results here.

\begin{figure}
    \centering
    \includegraphics[width=0.75\linewidth]{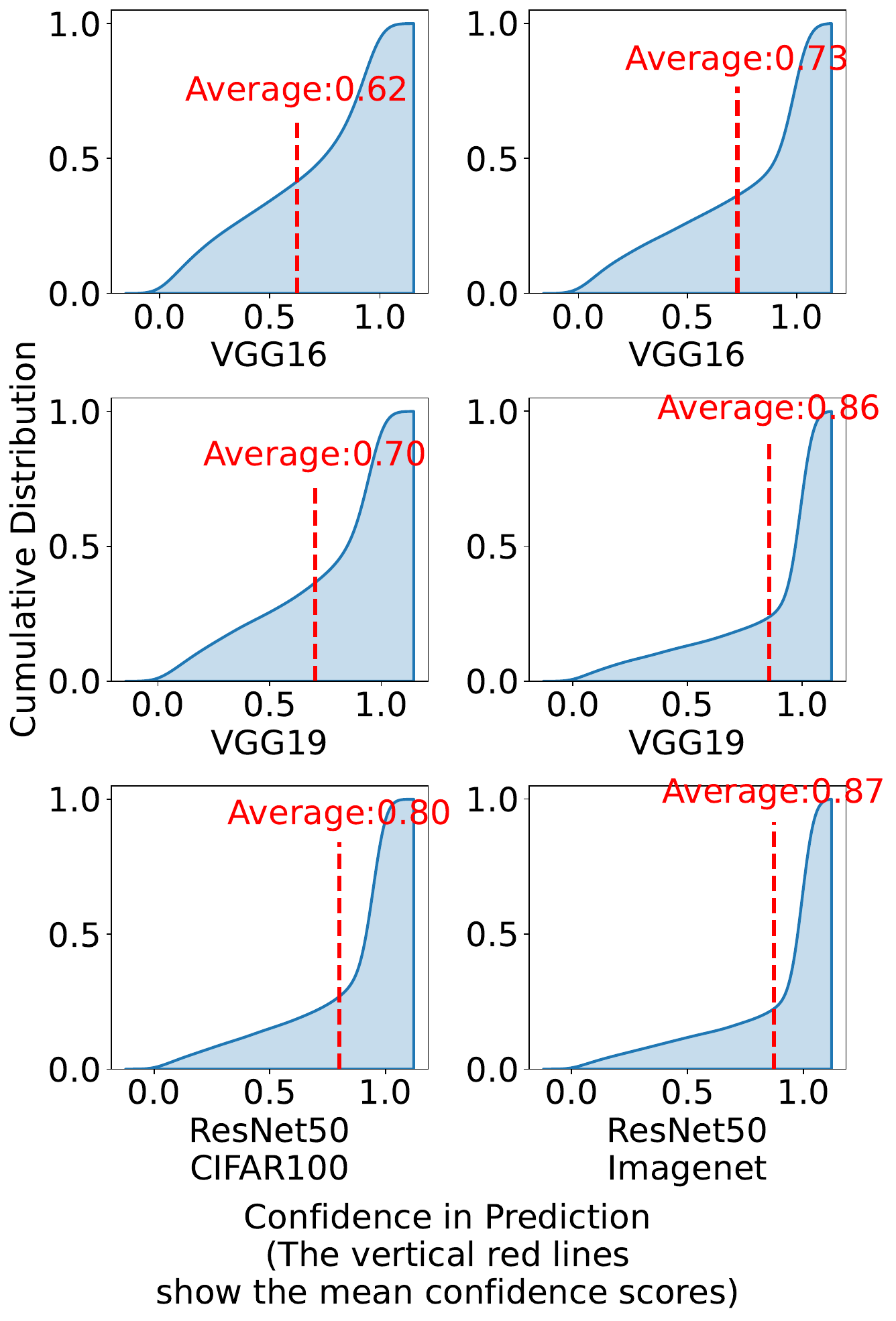}
    \caption{Cumulative distribution of the confidence values of \FW for the VGG16, VGG19, and ResNet50 \dnns for CIFAR100 (left) and ImageNet (right) benchmarks respectively.}
    \label{fig:model_conf_vals} \vspace{-0.5cm}
\end{figure}

\subsection{Impact of Confidence Threshold}

We first evaluate the impact of the confidence threshold $\alpha$.~The top row of Figure \ref{fig:model_conf_vals_acc_inference} shows the decrease in accuracy and the relative latency with respect to the original \dnn as a function of $\alpha$. As expected, increasing $\alpha$ increases the accuracy while also decreasing the gain in latency. As such, the confidence threshold $\alpha$ acts as a hyperparameter to find the needed trade-off between accuracy and latency. We notice that with VGG19, the overall accuracy actually increases by up to 0.49\%  for $0.4 \leq \alpha \leq 0.9$. In the best case, \FW reduces the inference time by up to 35\%, 29\% and 15\% with only 0.17\%, 3.75\%, and 6.75\% accuracy loss for VGG19, VGG16, and ResNet50 respectively. For the ImageNet benchmark, we observe that \FW achieves 18\%, 22\%, and 9\% less inference time with an accuracy degradation of 3\%, 2.5\%, and 6\% respectively. This effect may be due to many filters being shared among the semantic clusters creating polysemantic neurons \citep{olah2017feature}. 
As a result, the gain from such partitioning is smaller than the same for datasets like CIFAR100. This gain in performance depends on the frequency of activation of the semantic subgraphs. As such, we analyze the cumulative distribution of the confidence values in Figure \ref{fig:model_conf_vals} with CIFAR100 (top) and ImageNet (bottom) datasets. Figure \ref{fig:model_conf_vals}, shows that for both CIFAR100 and ImageNet datasets, we obtain high confidence value. The distribution of confidence values is skewed to the right with a high mean, which translates to a high frequency of activation of the subgraphs. \vspace{-0.1cm}

\subsection{DCS vs Existing Discriminative Metrics}
To evaluate the effectiveness of \gls{dcs} with respect to prior approaches, we use the metrics proposed in existing work while keeping the same inference structure of \FW. Figure \ref{fig:comp_acc} compares \gls{dcs} against gradient-based approaches \textit{Sensitivity} by \citep{mittal2019}  and \textit{Taylor} by \citep{molchanov2019taylor}, sparsity of activation based approach \textit{APOZ} by \citep{apoz2016}, channel-independence based approach \textit{CHIP} by \citep{sui2021chip}, and an approach based on channel importance named \textit{HRANK} by \citep{lin2020hrank}. All the approaches are compared without retraining the \dnn. Figure \ref{fig:comp_acc} shows that in the best case, \gls{dcs} has 15\% higher accuracy than the second-best approach \emph{Taylor} for VGG16 with 75\% sparsity (i.e., percentage of parameters dropped). For VGG19, DCS achieves in the best case 6.54\% higher accuracy than the second-best approach \emph{Taylor} at 63\% sparsity. Lastly, in the case of ResNet50, the best case is attained at 51\% sparsity, where DCS presents 14.87\% more accuracy than the second-best approach \emph{Sensitivity}. On average, \FW achieves 3.65\%, 4.25\%, and 2.36\% better accuracy than the second-best approaches for VGG16, VGG19, and ResNet50 respectively on CIFAR100 dataset. For ImageNet dataset, \FW  outperforms the second best approaches by 8.9\%,5.8\%, and 5.2\% on average for VGG16, VGG19, and ResNet50.

\vspace{-0.2cm} 

\begin{figure}[!h]
    \centering
    \includegraphics[width=1.0\linewidth, angle=270]{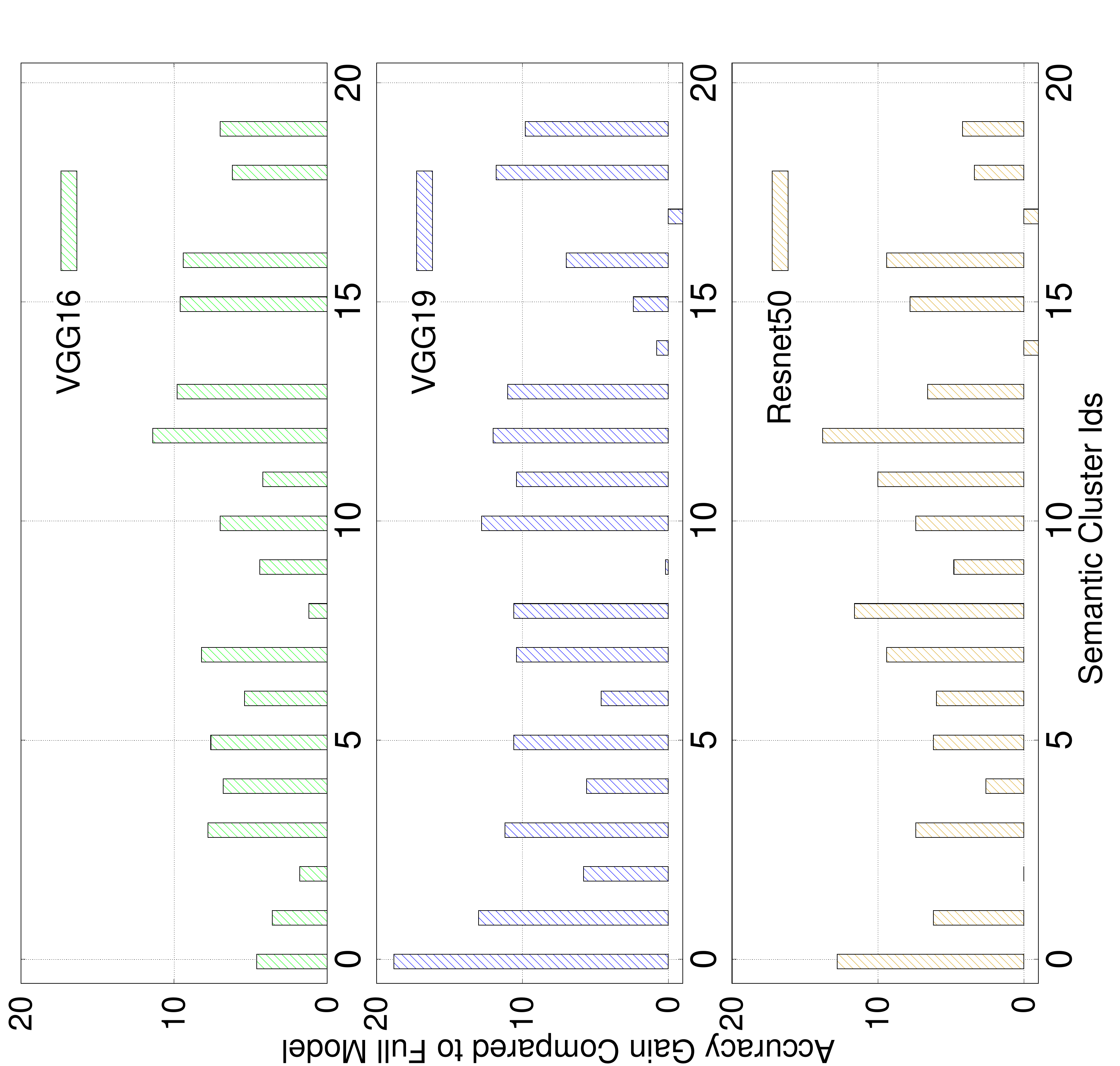}
    \caption{Performance gain compared to the original \gls{dnn}. \vspace{-0.5cm}}
    \label{fig:compare_to_orig} 
\end{figure}

\subsection{Visualization of Discriminative Features from Subgraphs}
\label{sec:tsne}
\begin{figure}[!h]
    \centering
    \includegraphics[width=0.9\linewidth, angle=270]{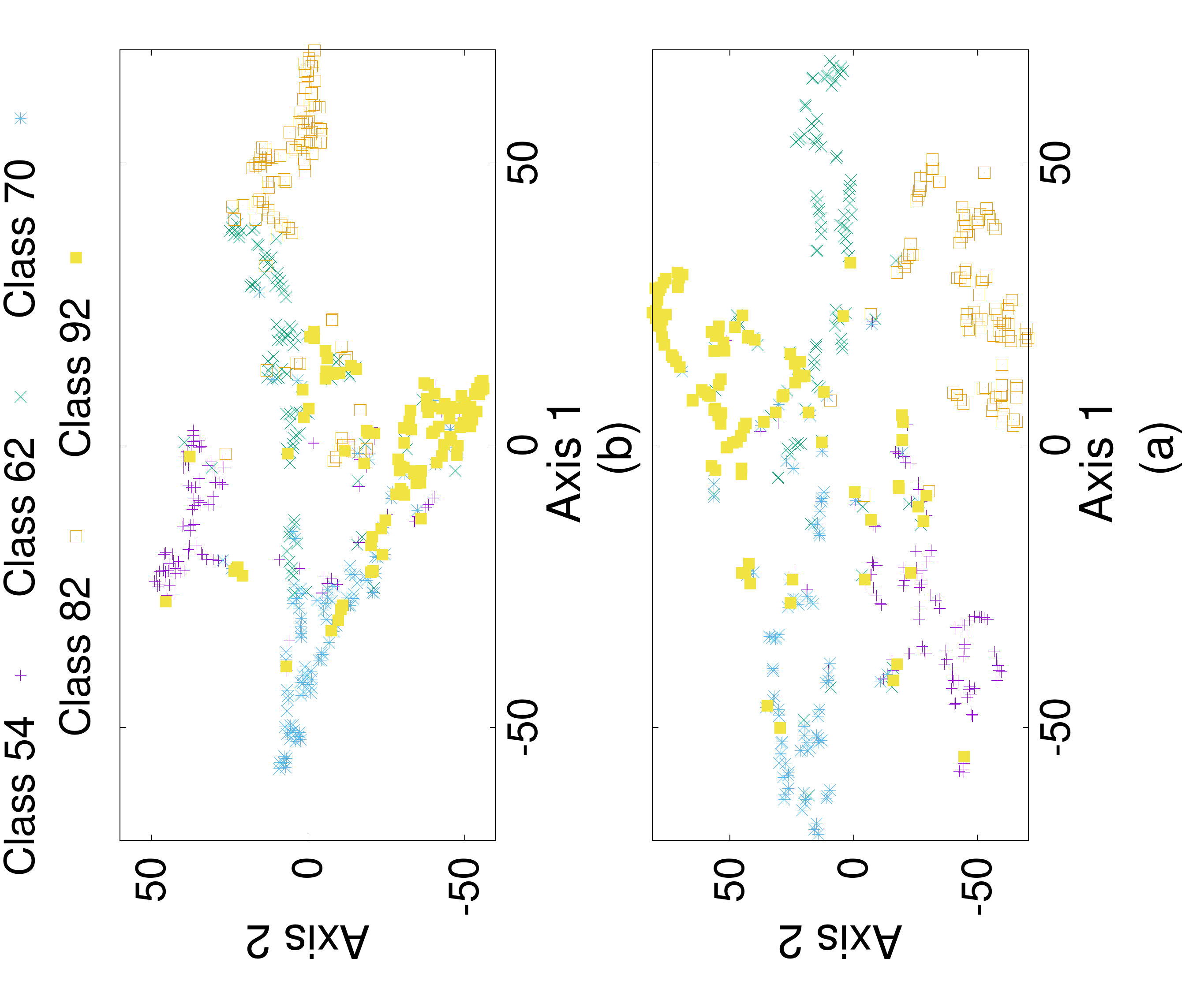}
    \caption{Comparison of the original model (a) and extracted subgraph (b) using TSNE visualization of the extracted features.\vspace{-0.5cm}}
    \label{fig:tsne}
\end{figure}

To understand if the extracted subgraphs have sufficient discriminative capability, we utilize the feature map generated from the convolution layer backbone of the \gls{dnn}. Our objective is to evaluate whether the features are sufficiently separated from each other. We plot the t-distributed stochastic neighbor embedding (t-SNE) visualization of the features extracted by VGG16 for the "flowers" cluster of CIFAR100. For comparison, we plot the t-SNE visualization for the original \gls{dnn} in Figure \ref{fig:tsne} (a). From Figure \ref{fig:tsne} (b), we observe that the feature maps are sufficiently separated to obtain good performance.

\subsection{Per-Cluster Accuracy Gain}
\label{supp:per_cluster_gain}
We ask the question: ``\textit{Can \FW perform better than the original \dnn when considering the accuracy obtained in individual clusters?}''. Figure \ref{fig:compare_to_orig} shows the accuracy gain obtained on the individual clusters by those subgraphs as compared to the original VGG16, VGG19, and ResNet50 \dnns.  In these experiments, we find the subgraphs with the lowest percentage of parameters retained while satisfying the constraint on the evaluation criterion posed in Equation \ref{eq:problem_formulation}.  Intriguingly, \FW provides on the average 5.73\%, 8.38\% and 6.36\% better per-cluster accuracy than the original VGG16, VGG19, and ResNet50 \glspl{dnn}, respectively, notwithstanding that the number of parameters have been reduced by 30\%, 50\%, and 44\%.~We believe the reason behind this improvement is that the semantic partitioning performed by \FW improves the \dnn explainability saving the \dnn from being ``less confused'' among different semantic clusters, which justifies better results when considering per-cluster accuracy. \vspace{-0.2cm}

    
     


\subsection{Latency and Energy Consumption on Mobile Devices}
\begin{figure}[!t]
    \centering
    \includegraphics[width=0.8\linewidth, angle=0]{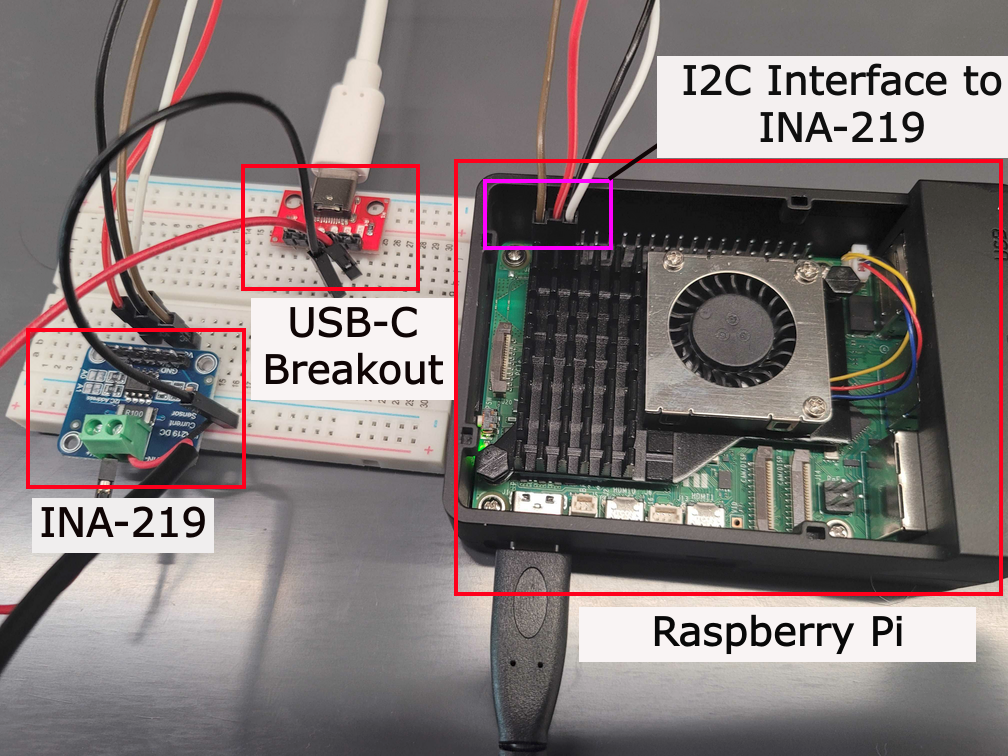}
    \caption{The setup for power measurement.}
    \label{fig:exp-setup-cv} \vspace{-0.5cm}
\end{figure}

We considered Raspberry-Pi-5 and Nvidia Jetson-Nano as example devices for running mobile \gls{cv} applications. Our experimental setup is shown in Figure \ref{fig:exp-setup-cv}. The Raspberry Pi runs a quad-core ARM A76 SoC running at upto 2.4 GHz with 8 GB LPDDR4 memory. In addition to being powered by a quad-core ARM Cortex A57 CPU, Jetson-Nano is equipped with 128 core Maxwell GPU. \smallskip

\begin{figure}[!h]
    \centering
    \includegraphics[width=0.45\linewidth, angle=270]{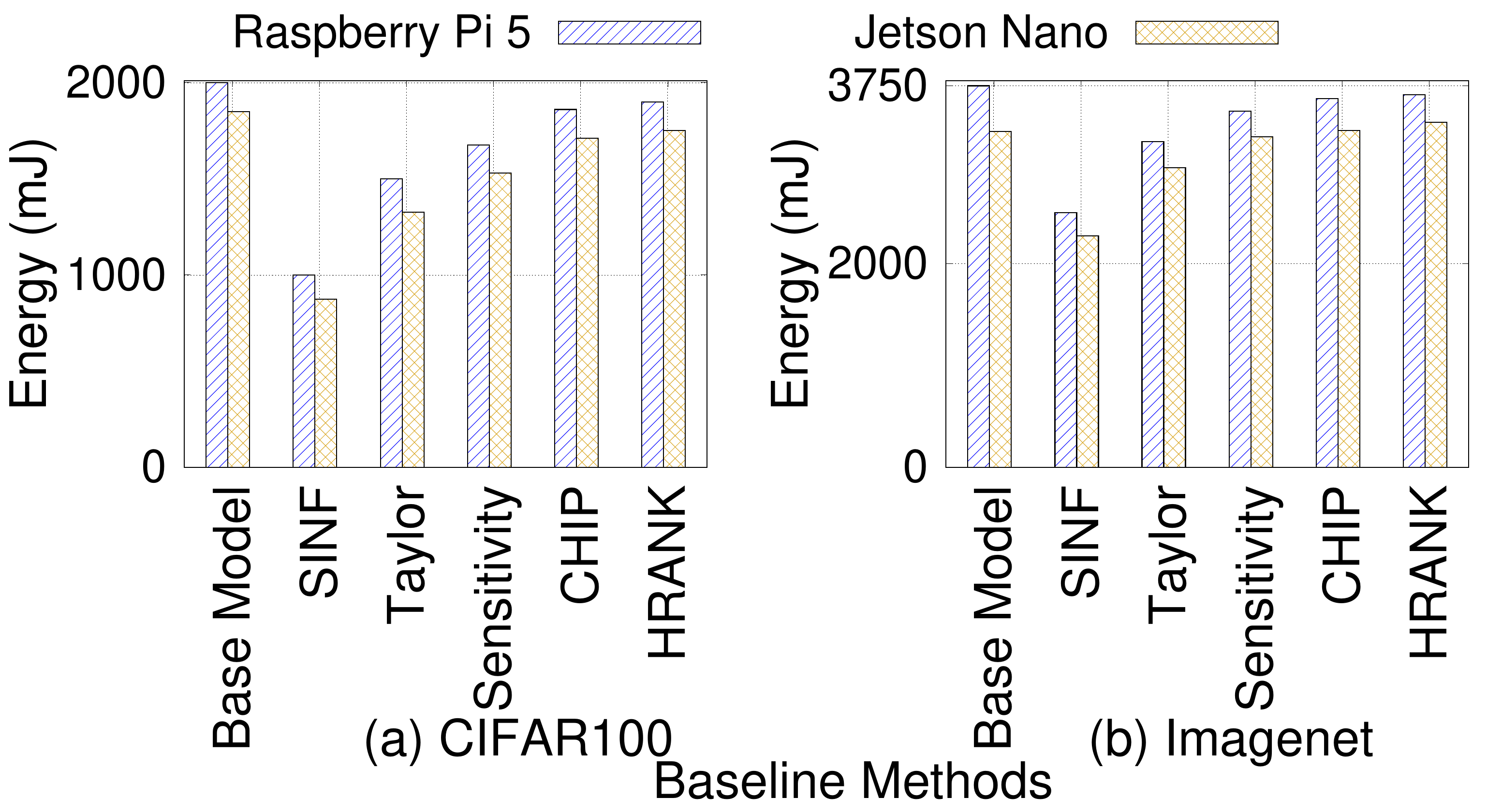}
    \caption{Power, \FW vs baselines.}
    \label{fig:power_measurement} \vspace{-0.8cm}
\end{figure}

\noindent \textbf{Energy Consumption:} We measure the power consumption of \FW and compare against the power consumption of the considered baselines. From \ref{fig:power_measurement}, we show that \FW outperforms the other methods in terms of power efficiency. Specifically, for Raspberry Pi 5 and Jetson Nano devices \FW respectively achieves 50\% and 52\% better energy efficiency on CIFAR100 benchmark and about 21\% and 22\% better energy efficiency on the ImageNet benchmark as compared to the second best approach \textit{Taylor}. When comparing with the base (static) model, \FW uses about 50\% less power for CIFAR100 and 38\% less power for ImageNet benchmark. 


\begin{table*}[!h]
    \centering
    \small
    \begin{tabular}{|c c c c c c|}
         \hline
         \textbf{\dnn} & \textbf{Dataset} & \textbf{Pruned} & \textbf{Criterion} & \textbf{Accuracy Loss} & \textbf{Difference} \\
         \hline
         \multirow{2}{*}{VGG16} & \multirow{2}{*}{CIFAR100} & 40.2\% & IterTVSPrune & 18.59\% & +9.75\% Accuracy\\

         &  & {43.8\%} & {DCS} & {8.84\%} & -3.6\% Parameters\\

         \hline

         \multirow{2}{*}{VGG16} & \multirow{2}{*}{CIFAR10} & 37.6\% & IterTVSPrune & 1.9\% & +0.61\% Accuracy\\
         &  & {42\%} & {DCS} & {1.29\%} & -4.4\% Parameters\\

         \hline

        \multirow{2}{*}{VGG19} & \multirow{2}{*}{CIFAR100} & 59\% & IterTVSPrune & 5.2\% & +0.05\% Accuracy\\

         &  &  59\% & {DCS} & {5.15\%} & +0\% Parameters \\
         
         \hline

         \multirow{2}{*}{VGG19} & \multirow{2}{*}{CIFAR10} & 49\% & IterTVSPrune & 1.3\% & +0.4\% Accuracy \\
         &  &  {49.65\%} &  {DCS} & {0.9\%} & -0.65\% Parameters\\

         \hline

         \multirow{2}{*}{ResNet50} & \multirow{2}{*}{CIFAR10}  & 34.1\% & IterTVSPrune & 9.94\% & +8.13\% Accuracy\\

         &  & {39.92\%} & {DCS} & {1.81\%} & -5.82\% Parameters \\

         \hline

         \multirow{2}{*}{ResNet50} & \multirow{2}{*}{ImageNet}  & 9.98\% & IterTVSPrune & 10.21\% & +2.41\% Accuracy\\

         &  & {12\%} & {DCS} & {7.8\%} & -2.02\% Parameters \\

         \hline      
    \end{tabular}
    \caption{Using DCS as a Pruning Criterion vs \textit{IterTVSPrune} (ICLR 2023). \vspace{-0.5cm}}
    \label{tab:pruning_performance}
\end{table*} 

\noindent \textbf{Inference Latency:}~We consider the time introduced by the common feature extractor (static part), semantic route predictor, and the semantic cluster specific subgraph (dynamic part). Our experiment shows that on an average, \FW achieves the lowest inference latency compared to the baselines which includes the base model itself. Specifically, \FW takes about 27\% and 24\% less time than the base model for the CIFAR100 benchmark while running on the Raspberry Pi 5 and Jetson Nano respectively. For the ImageNet Benchmark, \FW takes about 18\% and 16\% less inference time on Raspberry Pi 5 and Jetson Nano respectively than the base model itself. When considering the task adaptive deployment setting, we measure the latency of communication. We observe that sending the semantic subgraph reduces the average communication latency about 52\% than sending the base model for CIFAR100 benchmark. For the ImageNet benchmark, the communication latency is reduced by about 30\%. 

\subsection{DCS as Pruning Criterion}
Viewing the dataset $\mathcal{D}$ as a single macro-cluster DCS can be applied to determine the most relevant filters effectively acting as a pruning criterion. For comparison, we consider the state-of-the-art \textit{IterTVSPrune} by \citep{murti2023tvsprune} published at ICLR 2023, which also does not require fine-tuning. For a fair comparison, we have taken the percentage of parameters pruned by \textit{IterTVSPrune} at each layer and set the same pruning threshold for DCS. \ref{tab:pruning_performance} summarizes the performance achieved by \textit{DCS} and \textit{IterTVSPrune}. We did not compare performance on CIFAR100 with ResNet50 as the authors of \citep{murti2023tvsprune} did not provide the performance of their approach on ResNet50 trained with CIFAR100. We notice that for different \gls{dnn} structures and datasets, \gls{dcs} achieves substantially better performance in three out of five settings considered while achieving similar performance in the remaining two settings. For VGG19, both our technique and the IterTVSPrune have pruned a significant amount of weights – respectively about 50\% and 60\% for CIFAR10 and CIFAR100 – possibly causing the DNN to reach a lower bound on its predictive capability thereby causing similar performance of both techniques. The best results are obtained in the case of VGG16 trained on CIFAR100 and ResNet50 trained on CIFAR10, where we see respectively 9.75\% and 8.13\% accuracy gain with 3.6\% and 5.82\% less parameters. On average, DCS presents 3.78\% better accuracy and 2.89\% less parameters than \textit{IterTVSPrune}. 

\begin{figure}
    \centering
    \includegraphics[width=0.45\linewidth, angle=270]{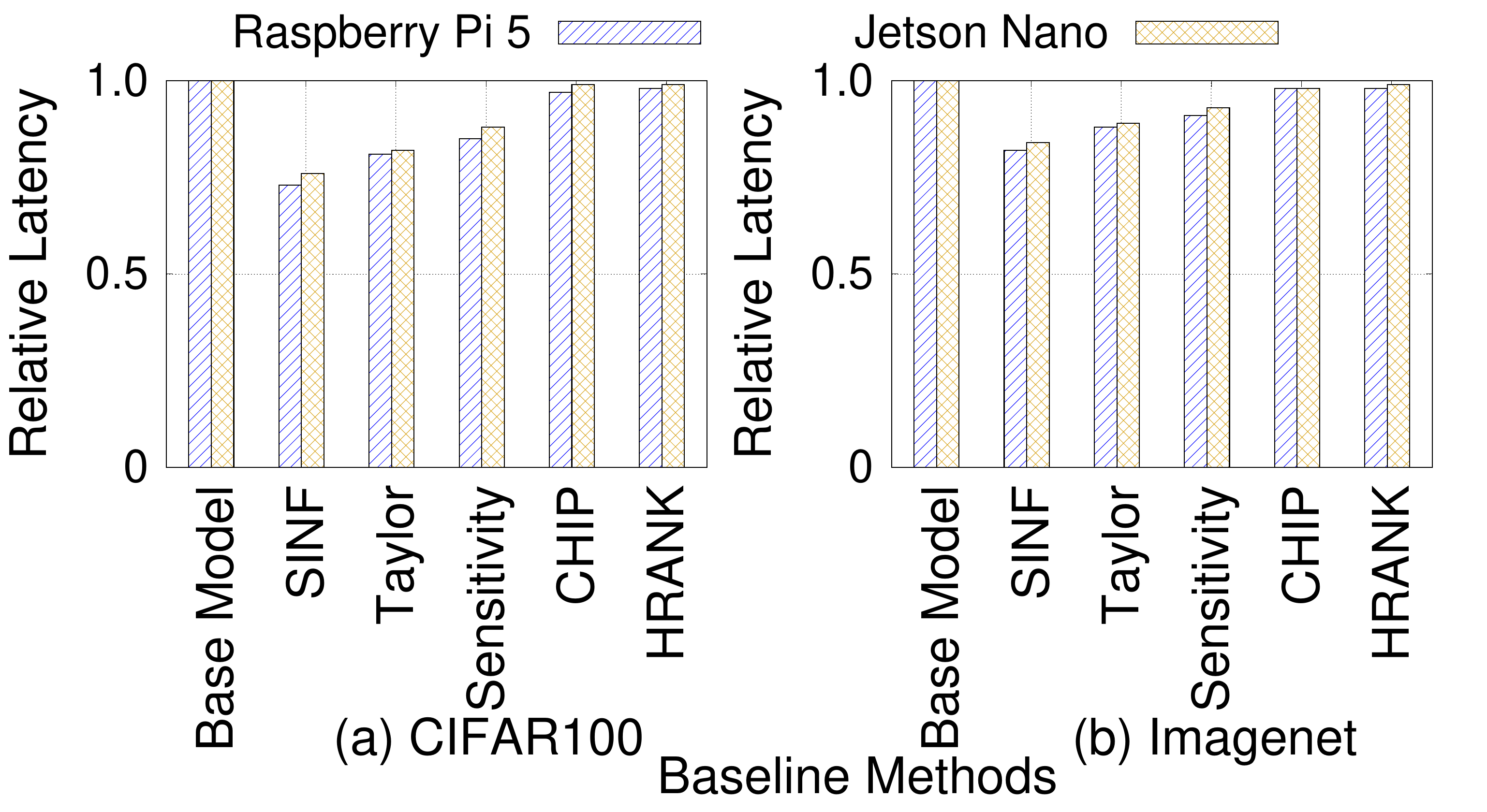}
    \caption{Latency, \FW vs baselines.}
    \label{fig:latency_measurement} \vspace{-0.8cm}
\end{figure}

\section{Concluding Remarks}\label{sec:conclusions}

In this paper, we propose \textit{Semantic Inference} (\FW), a framework for accelerating \dnns without compromising accuracy. Central to \FW is the \emph{Discriminative Capability Score} (DCS), which we employ to identify subgraphs within large \dnns that discriminate specific semantic clusters. Unlike model compression (e.g., pruning, quantization), \FW requires no fine-tuning and operates at the cluster level. We have benchmarked \FW on VGG16, VGG19, and ResNet50 using CIFAR100 and ImageNet, showing promising performance as compared to the existing methods and proving its efficiency for edge deployment.

\section*{Acknowledgement}

This work has been funded in part by the National Science Foundation under grant CNS-2312875, by the Air Force Office of Scientific Research under contract number FA9550-23-1-0261, by the Office of Naval Research under award number N00014-23-1-2221.

\section*{Impact Statement}

This paper presents work whose goal is to advance the field of 
Machine Learning. There are many potential societal consequences 
of our work, none which we feel must be specifically highlighted here.

\nocite{langley00}

\bibliography{main, egbib, bib-francesco}

\begin{thebibliography}{34}
\providecommand{\natexlab}[1]{#1}
\providecommand{\url}[1]{\texttt{#1}}
\expandafter\ifx\csname urlstyle\endcsname\relax
  \providecommand{\doi}[1]{doi: #1}\else
  \providecommand{\doi}{doi: \begingroup \urlstyle{rm}\Url}\fi

\bibitem[Ao~Wang(2024)]{THU-MIGyolov10}
Ao~Wang, Hui~Chen, L. L. e.~a.
\newblock Yolov10: Real-time end-to-end object detection.
\newblock \emph{arXiv preprint arXiv:2405.14458}, 2024.

\bibitem[Bau et~al.(2017)Bau, Zhou, Khosla, Oliva, and Torralba]{bau2017networkdissection}
Bau, D., Zhou, B., Khosla, A., Oliva, A., and Torralba, A.
\newblock Network dissection: Quantifying interpretability of deep visual representations.
\newblock In \emph{Proceedings of the IEEE conference on computer vision and pattern recognition}, pp.\  6541--6549, 2017.

\bibitem[Cai et~al.(2020)Cai, Yao, Dong, Gholami, Mahoney, and Keutzer]{cai2020zeroq}
Cai, Y., Yao, Z., Dong, Z., Gholami, A., Mahoney, M.~W., and Keutzer, K.
\newblock Zeroq: A novel zero shot quantization framework.
\newblock In \emph{Proceedings of the IEEE/CVF Conference on Computer Vision and Pattern Recognition}, pp.\  13169--13178, 2020.

\bibitem[Chen et~al.(2023)Chen, Ma, Fang, Zheng, Yu, and Tian]{chen2023a}
Chen, Y., Ma, Z., Fang, W., Zheng, X., Yu, Z., and Tian, Y.
\newblock A unified framework for soft threshold pruning.
\newblock In \emph{The Eleventh International Conference on Learning Representations}, 2023.
\newblock URL \url{https://openreview.net/forum?id=cCFqcrq0d8}.

\bibitem[Deng et~al.(2009)Deng, Dong, Socher, Li, Li, and Fei-Fei]{deng2009imagenet}
Deng, J., Dong, W., Socher, R., Li, L.-J., Li, K., and Fei-Fei, L.
\newblock Imagenet: A large-scale hierarchical image database.
\newblock In \emph{2009 IEEE conference on computer vision and pattern recognition}, pp.\  248--255. Ieee, 2009.

\bibitem[Dong et~al.(2022)Dong, Mao, and Zhang]{Dong2022EP}
Dong, R., Mao, Y., and Zhang, J.
\newblock Resource-constrained edge ai with early exit prediction.
\newblock \emph{Journal of Communications and Information Networks}, 7\penalty0 (2):\penalty0 122--134, 2022.
\newblock \doi{10.23919/JCIN.2022.9815196}.

\bibitem[Gajjala et~al.(2020)Gajjala, Banchhor, Abdelmoniem, Dutta, Canini, and Kalnis]{gajjala2020}
Gajjala, R.~R., Banchhor, S., Abdelmoniem, A.~M., Dutta, A., Canini, M., and Kalnis, P.
\newblock Huffman coding based encoding techniques for fast distributed deep learning.
\newblock In \emph{Proceedings of the 1st Workshop on Distributed Machine Learning}, DistributedML'20, pp.\  21–27, New York, NY, USA, 2020. Association for Computing Machinery.
\newblock ISBN 9781450381826.
\newblock \doi{10.1145/3426745.3431334}.
\newblock URL \url{https://doi.org/10.1145/3426745.3431334}.

\bibitem[Garg \& Moschitti(2021)Garg and Moschitti]{garg2021will}
Garg, S. and Moschitti, A.
\newblock Will this question be answered? question filtering via answer model distillation for efficient question answering.
\newblock In \emph{Proceedings of the 2021 Conference on Empirical Methods in Natural Language Processing}, pp.\  7329--7346, 2021.

\bibitem[Han et~al.(2023)Han, Park, Ham, Lee, and Moon]{han2023blockloss}
Han, D.-J., Park, J., Ham, S., Lee, N., and Moon, J.
\newblock Improving low-latency predictions in multi-exit neural networks via block-dependent losses.
\newblock \emph{IEEE Transactions on Neural Networks and Learning Systems}, pp.\  1--9, 2023.
\newblock \doi{10.1109/TNNLS.2023.3282249}.

\bibitem[Han et~al.(2015{\natexlab{a}})Han, Mao, and Dally]{han2015deep}
Han, S., Mao, H., and Dally, W.~J.
\newblock Deep compression: Compressing deep neural networks with pruning, trained quantization and huffman coding.
\newblock \emph{arXiv preprint arXiv:1510.00149}, 2015{\natexlab{a}}.

\bibitem[Han et~al.(2015{\natexlab{b}})Han, Pool, Tran, and Dally]{han2015learning}
Han, S., Pool, J., Tran, J., and Dally, W.~J.
\newblock Learning both weights and connections for efficient neural networks.
\newblock In \emph{Proceedings of the 28th International Conference on Neural Information Processing Systems-Volume 1}, pp.\  1135--1143, 2015{\natexlab{b}}.

\bibitem[He et~al.(2016)He, Zhang, Ren, and Sun]{he2016deep}
He, K., Zhang, X., Ren, S., and Sun, J.
\newblock {Deep Residual Learning for Image Recognition}.
\newblock In \emph{Proceedings of the IEEE Conference on Computer Vision and Pattern Recognition (CVPR)}, pp.\  770--778, 2016.

\bibitem[Hu et~al.(2016)Hu, Peng, Tai, and Tang]{apoz2016}
Hu, H., Peng, R., Tai, Y., and Tang, C.
\newblock Network trimming: {A} data-driven neuron pruning approach towards efficient deep architectures.
\newblock \emph{CoRR}, abs/1607.03250, 2016.
\newblock URL \url{http://arxiv.org/abs/1607.03250}.

\bibitem[Krizhevsky \& Hinton(2009)Krizhevsky and Hinton]{krizhevsky2009learning}
Krizhevsky, A. and Hinton, G.
\newblock {Learning Multiple Layers of Features from Tiny Images}.
\newblock 2009.

\bibitem[Li et~al.(2017)Li, Kadav, Durdanovic, Samet, and Graf]{li2017pruning}
Li, H., Kadav, A., Durdanovic, I., Samet, H., and Graf, H.~P.
\newblock Pruning filters for efficient convnets.
\newblock In \emph{International Conference on Learning Representations}, 2017.
\newblock URL \url{https://openreview.net/forum?id=rJqFGTslg}.

\bibitem[Li et~al.(2020)Li, Yan, Lin, Zheng, Li, Zhang, Yang, and Ji]{li2020pams}
Li, H., Yan, C., Lin, S., Zheng, X., Li, Y., Zhang, B., Yang, F., and Ji, R.
\newblock Pams: Quantized super-resolution via parameterized max scale.
\newblock \emph{arXiv preprint arXiv:2011.04212}, 2020.

\bibitem[Lin et~al.(2020)Lin, Ji, Wang, Zhang, Zhang, Tian, and Shao]{lin2020hrank}
Lin, M., Ji, R., Wang, Y., Zhang, Y., Zhang, B., Tian, Y., and Shao, L.
\newblock Hrank: Filter pruning using high-rank feature map.
\newblock In \emph{Proceedings of the IEEE/CVF Conference on Computer Vision and Pattern Recognition (CVPR)}, pp.\  1529--1538, 2020.

\bibitem[Matsubara et~al.(2021)Matsubara, Levorato, and Restuccia]{matsubara2021split}
Matsubara, Y., Levorato, M., and Restuccia, F.
\newblock {Split Computing and Early Exiting for Deep Learning Applications: Survey and Research Challenges}.
\newblock \emph{ACM Computing Surveys (CSUR)}, 2021.

\bibitem[Mittal et~al.(2019)Mittal, Bhardwaj, Khapra, and Ravindran]{mittal2019}
Mittal, D., Bhardwaj, S., Khapra, M.~M., and Ravindran, B.
\newblock Studying the plasticity in deep convolutional neural networks using random pruning.
\newblock \emph{Machine Vision and Applications}, 30\penalty0 (2):\penalty0 203--216, 2019.

\bibitem[Molchanov et~al.(2019)Molchanov, Mallya, Tyree, Frosio, and Kautz]{molchanov2019taylor}
Molchanov, P., Mallya, A., Tyree, S., Frosio, I., and Kautz, J.
\newblock Importance estimation for neural network pruning.
\newblock In \emph{Proceedings of the IEEE Conference on Computer Vision and Pattern Recognition}, 2019.

\bibitem[Murti et~al.(2023)Murti, Narshana, and Bhattacharyya]{murti2023tvsprune}
Murti, C., Narshana, T., and Bhattacharyya, C.
\newblock {TVSP}rune - pruning non-discriminative filters via total variation separability of intermediate representations without fine tuning.
\newblock In \emph{The Eleventh International Conference on Learning Representations}, 2023.
\newblock URL \url{https://openreview.net/forum?id=sZI1Oj9KBKy}.

\bibitem[Narayan et~al.(2023)Narayan, Hanawal, and Bhardwaj]{Hari2023unsupervisedEE}
Narayan, H., Hanawal, M.~K., and Bhardwaj, A.
\newblock Unsupervised early exit in dnns with multiple exits.
\newblock In \emph{Proceedings of the Second International Conference on AI-ML Systems}, AIMLSystems '22, New York, NY, USA, 2023. Association for Computing Machinery.
\newblock ISBN 9781450398473.
\newblock \doi{10.1145/3564121.3564137}.
\newblock URL \url{https://doi.org/10.1145/3564121.3564137}.

\bibitem[Olah et~al.(2017)Olah, Mordvintsev, and Schubert]{olah2017feature}
Olah, C., Mordvintsev, A., and Schubert, L.
\newblock Feature visualization.
\newblock \emph{Distill}, 2017.
\newblock \doi{10.23915/distill.00007}.
\newblock https://distill.pub/2017/feature-visualization.

\bibitem[Park et~al.(2015)Park, Kim, Kim, Kim, Kim, Yoon, and Yoo]{BLDNN2015}
Park, E., Kim, D., Kim, S., Kim, Y.-D., Kim, G., Yoon, S., and Yoo, S.
\newblock Big/little deep neural network for ultra low power inference.
\newblock In \emph{2015 International Conference on Hardware/Software Codesign and System Synthesis (CODES+ISSS)}, pp.\  124--132, 2015.
\newblock \doi{10.1109/CODESISSS.2015.7331375}.

\bibitem[Phuong \& Lampert(2019)Phuong and Lampert]{Phuong2019DistEE}
Phuong, M. and Lampert, C.
\newblock Distillation-based training for multi-exit architectures.
\newblock In \emph{2019 IEEE/CVF International Conference on Computer Vision (ICCV)}, pp.\  1355--1364, 2019.
\newblock \doi{10.1109/ICCV.2019.00144}.

\bibitem[Pomponi et~al.(2022)Pomponi, Scardapane, and Uncini]{pomponi2022probabilistic}
Pomponi, J., Scardapane, S., and Uncini, A.
\newblock A probabilistic re-intepretation of confidence scores in multi-exit models.
\newblock \emph{Entropy}, 24\penalty0 (1), 2022.
\newblock ISSN 1099-4300.
\newblock \doi{10.3390/e24010001}.
\newblock URL \url{https://www.mdpi.com/1099-4300/24/1/1}.

\bibitem[Qin et~al.(2022)Qin, Ding, Zhang, YAN, Liu, Dang, Liu, and Liu]{qin2022bibert}
Qin, H., Ding, Y., Zhang, M., YAN, Q., Liu, A., Dang, Q., Liu, Z., and Liu, X.
\newblock Bi{BERT}: Accurate fully binarized {BERT}.
\newblock In \emph{International Conference on Learning Representations}, 2022.
\newblock URL \url{https://openreview.net/forum?id=5xEgrl_5FAJ}.

\bibitem[Simonyan \& Zisserman(2014)Simonyan and Zisserman]{vgg2014}
Simonyan, K. and Zisserman, A.
\newblock Very deep convolutional networks for large-scale image recognition.
\newblock \emph{arXiv preprint arXiv:1409.1556}, 2014.

\bibitem[Sui et~al.(2021)Sui, Yin, Xie, Phan, Zonouz, and Yuan]{sui2021chip}
Sui, Y., Yin, M., Xie, Y., Phan, H., Zonouz, S.~A., and Yuan, B.
\newblock {CHIP}: {CH}annel independence-based pruning for compact neural networks.
\newblock In Beygelzimer, A., Dauphin, Y., Liang, P., and Vaughan, J.~W. (eds.), \emph{Advances in Neural Information Processing Systems}, 2021.
\newblock URL \url{https://openreview.net/forum?id=EmeWbcWORRg}.

\bibitem[Teerapittayanon et~al.()Teerapittayanon, McDanel, and Kung]{teerapittayanonbranchynet}
Teerapittayanon, S., McDanel, B., and Kung, H.
\newblock Branchynet: Fast inference via early exiting from deep neural networks.

\bibitem[Tu et~al.(2023)Tu, Hu, Chen, and Wang]{tu2023toward}
Tu, Z., Hu, J., Chen, H., and Wang, Y.
\newblock Toward accurate post-training quantization for image super resolution.
\newblock In \emph{Proceedings of the IEEE/CVF Conference on Computer Vision and Pattern Recognition}, pp.\  5856--5865, 2023.

\bibitem[Verd{\'u}(2014)]{verdu2014total}
Verd{\'u}, S.
\newblock Total variation distance and the distribution of relative information.
\newblock In \emph{2014 Information Theory and Applications Workshop (ITA)}, pp.\  1--3. IEEE, 2014.

\bibitem[Wang et~al.(2019)Wang, Mo, Lin, Wang, and Du]{Wang2019DyNExit}
Wang, M., Mo, J., Lin, J., Wang, Z., and Du, L.
\newblock Dynexit: A dynamic early-exit strategy for deep residual networks.
\newblock In \emph{2019 IEEE International Workshop on Signal Processing Systems (SiPS)}, pp.\  178--183, 2019.
\newblock \doi{10.1109/SiPS47522.2019.9020551}.

\bibitem[Zhong et~al.(2022)Zhong, Lin, Li, Li, Shen, Chao, Wu, and Ji]{Zhong2022}
Zhong, Y., Lin, M., Li, X., Li, K., Shen, Y., Chao, F., Wu, Y., and Ji, R.
\newblock Dynamic dual trainable bounds for ultra-low precision super-resolution networks.
\newblock In \emph{Computer Vision – ECCV 2022: 17th European Conference, Tel Aviv, Israel, October 23–27, 2022, Proceedings, Part XVIII}, pp.\  1–18, Berlin, Heidelberg, 2022. Springer-Verlag.
\newblock ISBN 978-3-031-19796-3.
\newblock \doi{10.1007/978-3-031-19797-0_1}.
\newblock URL \url{https://doi.org/10.1007/978-3-031-19797-0_1}.

\end{thebibliography}
\bibliographystyle{icml2025}

\newpage
\appendix
\section{Details of the Experiment Motivating \FW}
\label{sec:motivating_exp_1}
To obtain the results in the top portion of Figure \ref{fig:activation_pattern}, we have observed the activation pattern of the filters of a specific layer of the \gls{dnn} for a given class. To obtain the activation pattern of the filters of layer $L$ for a specific class $c$, we feed the input samples of that class and obtain the activation vectors $A_L \in \mathbb{R}^{N_c \times H \times W}$ from layer $L$, where $N_c, H, W$ are number of channels, height and width of the activation. Next, we take the mean value of the activations for each filter giving us activation vectors of length $N_c$ for each sample. For future reference, we will denote this with $\Tilde{A}_L$. Next, we take the expected values of the activation to obtain the filter activation pattern of layer $L$ for class $c$. For ease of visualization, we standardize the activation pattern using min-max standardization. 

To obtain the filter activation frequency in the bottom side of Figure \ref{fig:activation_pattern}, we take the $\Tilde{A}_L$ for each sample and stack them along the sample axis. This gives us a matrix of size $\mathbb{R}^{N_samples \times N_c}$. For each filter $f$ (each column represents a filter of layer $L$ now), we calculate the maximum value of activation $max_f$ and minimum value of activation $min_f$. We consider the filter to be activated if its activation value exceeds 70\% of the range $max_f - min_f$. Next, we obtain class-wise activation frequency $freq_c$ for each filter where $c$ denotes the class for which the activation frequency is being calculated. Next, we assign each filter with the top 20 classes for which it gets most activated. \smallskip





\section{DCS Algorithm}

\begin{algorithm}
\caption{DCS: Discriminative Capability Scoring for Filters in Layer $l$ and Semantic Cluster $\gamma$}
\label{alg:dcs}
\begin{algorithmic}[1]

\REQUIRE Dataset $\mathcal{D}_\gamma = \{(X^j, t^j)\}_{j=1}^{|\mathcal{D}_\gamma|}$ for semantic cluster $\gamma$
\REQUIRE Pretrained $L$-layer DNN $\mathcal{F} = \mathcal{F}_{L-1} \circ \ldots \circ \mathcal{F}_0$
\REQUIRE Objective function $\mathcal{L}_{DOF}$
\ENSURE Discriminative Capability Score $\text{DCS}^l$ for filters in layer $l$

\STATE Initialize score list $\mathbf{s}_l \gets []$

\FOR{each $(X^j, t^j)$ in $\mathcal{D}_\gamma$}
    \STATE $\mathbf{A}^j_l \gets \mathcal{F}_l \circ \ldots \circ \mathcal{F}_0(X^j)$
    \STATE $\tilde{\mathbf{A}}^j_l \gets \mathcal{P}(\mathbf{A}^j_l)$ \COMMENT{adaptive pooling to $k \times k$}
    \STATE $\mathbf{F}^j_l \gets \text{Flatten}(\tilde{\mathbf{A}}^j_l)$
\ENDFOR

\STATE $\mathbf{W}^*_l \gets \arg\min_{\mathbf{W}} \frac{1}{|\mathcal{D}_\gamma|} \sum_j \mathcal{L}_{DOF}(\mathbf{W} \cdot \mathbf{F}^j_l, t^j)$

\STATE $\mathbf{I}_l \gets \mathbf{W}^*_l \odot \nabla_{\mathbf{W}^*_l} \bar{\mathcal{L}}_{DOF}$

\FOR{$i = 0$ to $C^l_{out} - 1$}
    \STATE Append $\|\mathbf{I}_l[:, i]\|_2$ to $\mathbf{s}_l$
\ENDFOR

\STATE $\text{DCS}_i^l \gets \sqrt{\sum_j |s_j|}$ \COMMENT{$j$ indexes features of filter $i$}

\end{algorithmic}
\end{algorithm}


\section{Algorithm for Subgraph Extraction}
\label{sec:subgraph_extract}
\begin{algorithm}[!h]
\caption{Subgraph Extraction for Semantic Clusters}
\label{alg:subgraph_extraction}
\begin{algorithmic}[1]

\REQUIRE Partitioned Dataset $\mathcal{D} = \{\mathcal{D}_{\gamma_1}, \mathcal{D}_{\gamma_2}, \ldots, \mathcal{D}_{\gamma_K}\}$
\REQUIRE Pretrained DNN $\mathcal{F} = \mathcal{F}_{L-1} \circ \ldots \circ \mathcal{F}_0$
\REQUIRE Discriminative Objective Function $\mathcal{L}_{DOF}$
\REQUIRE Filter retention percentages $r_L$ at layer $L$, $r_M$ at layer $M$
\REQUIRE Accuracy threshold $\tau_{acc}$
\ENSURE Filter annotations $\mathbf{SA}$[] for extracted subgraphs

\STATE Initialize empty dictionary $\mathbf{SA} \gets \{\}$

\FOR{each $\mathcal{D}_{\gamma_i}$ in $\mathcal{D}$}
    \STATE $\mathbf{SA}_{\gamma_i} \gets \{\}$

    \FOR{$l = L$ to $M$ decreasing}
        \STATE $r_l \gets r_M + \frac{(l - M)(r_L - r_M)}{L - M}$ \COMMENT{Linear interpolation of retention rate}
        \STATE $\text{DCS}^l \gets DCS(\mathcal{F}, \mathcal{D}_{\gamma_i}, \mathcal{L}_{DOF})$
        \STATE Rank filters in layer $l$ by DCS$^l$
        \STATE Save indices of top $r_l\%$ filters in $\mathbf{SA}_{\gamma_i}[l]$
    \ENDFOR

    \STATE $acc_{avg} \gets \frac{1}{K} \sum_{i=1}^{K} \text{accuracy}(\mathcal{F}_{\gamma_i})$

    \IF{$acc_{avg} \geq \tau_{acc}$}
        \STATE Save $\mathbf{SA}_{\gamma_i}$ in $\mathbf{SA}[\gamma_i]$
    \ENDIF
\ENDFOR

\end{algorithmic}
\end{algorithm}

\section{Experimental Setup}
\label{sec:exp-set}

\noindent \textbf{Hyperparameters:}~We set $r_L$ between 90\% and 10\%, with steps of 10, while $r_{M}$ is set between 10\% and 1\%, with steps of 2. This way, we vary the number of retained filters in different layers allowing us to find multiple sub-graphs satisfying our constraint set discussed in Section 3. Based on the application-level performance constraint, we can choose the optimum model based on additional requirements (e.g. sub-graph size, latency). We linearly decrease the percentage of filters retained from layer $M$ to layer $L$  according to the Equation presented in Line 5 of Algorithm \ref{alg:subgraph_extraction}. Since we are considering classification tasks, categorical cross entropy is used for $\mathcal{L}_{DOF}$.\smallskip

\noindent \textbf{Dataset, Base DNNs, and Baselines.}~We have used the well-known CIFAR100 \citep{krizhevsky2009learning} and ImageNet \citep{deng2009imagenet} for image classification. CIFAR100 has 100 classes and the entire dataset is labeled into 20 super-classes corresponding to 20 coarse labels corresponding to our semantic clusters. With ImageNet we form 6 semantic clusters each consisting of 5 fine-grained classes. The semantic clusters formed are \textit{fish}, \textit{bird}, \textit{lizards}, \textit{animal}, \textit{insects}, and \textit{seafish}. A summary of the semantic clusters and their member classes are provided in Table \ref{tab:imagenet_cluster}. We have considered VGG16 and VGG19 \citep{vgg2014} as well as ResNet-50 \citep{he2016deep} for base DNNs. To the best of our knowledge, there is no prior work on semantic clustering. As such, we adapt pruning methods to use them without fine-tuning, i.e.,  \citep{molchanov2019taylor}, \citep{apoz2016}, \citep{mittal2019}, \citep{sui2021chip}, and \citep{lin2020hrank}. \vspace{-0.1cm}

\begin{table}[!h]
\resizebox{\linewidth}{!}{
\begin{tabular}{|c|c|}
\hline
Semantic Cluster & Class Label (Wordnet Label)                                                                                                               \\ \hline
Fishes           & \begin{tabular}[c]{@{}c@{}}2 (great white shark), 3 (tiger shark), \\4 (hammerhead shark), 5 (electric ray), \\6 (sting ray)\end{tabular}  \\ \hline
Birds            & \begin{tabular}[c]{@{}c@{}}10 (brambling), 11 (goldfinch), 12 (house finch), \\ 13 (junco), 14 (indigo bunting)\end{tabular}              \\ \hline
Lizards          & \begin{tabular}[c]{@{}c@{}}37 (box turtle), 38 (banded gecko), \\39 (common iguana), 40 (American chameleon), \\41 (whiptail)\end{tabular} \\ \hline
Animals          & \begin{tabular}[c]{@{}c@{}}253 (basenji), 261 (keeshond), 276 (hyena),\\  283 (persian cat), 298 (mongoose)\end{tabular}                  \\ \hline
Insects          & \begin{tabular}[c]{@{}c@{}}305 (dung beetle), 308 (fly), 309 (bee), \\ 310 (ant), 311 (grasshopper)\end{tabular}                          \\ \hline
Sea fish         & \begin{tabular}[c]{@{}c@{}}393 (anemone fish), 394 (sturgeon), 395 (gar),\\  396 (lionfish), 397 (puffer)\end{tabular}                    \\ \hline
\end{tabular}
}
\caption{Summary of the semantic clusters formed from ImageNet dataset.}
\label{tab:imagenet_cluster} \vspace{-0.5cm}
\end{table}



\end{document}